\documentclass[twoside,11pt]{article}
\usepackage{jmlr2e}

\jmlrheading{1}{2008}{1-10}{04/08}{04/08}{Gretton, Borgwardt, Rasch,
  Sch\"olkopf and Smola}

\ShortHeadings{A Kernel Method for the Two-Sample Problem}{Gretton,
  Borgwardt, Rasch, Sch\"olkopf and Smola} 

\firstpageno{1} 

\usepackage{amsmath}
\usepackage{amssymb}

\usepackage{eucal}

\usepackage[all]{xy}
\CompileMatrices

\newcommand{\eq}[1]{(\ref{#1})}

\newcommand{\inner}[2]{\left\langle #1,#2 \right\rangle}
\newcommand{\rbr}[1]{\left(#1\right)}
\newcommand{\sbr}[1]{\left[#1\right]}
\newcommand{\cbr}[1]{\left\{#1\right\}}
\newcommand{\nbr}[1]{\left\|#1\right\|}
\newcommand{\abr}[1]{\left|#1\right|}

\newcommand{\RR}{\mathbb{R}}

\newcommand{\one}{\mathbf{1}}

\newcommand{\Hcal}{\mathcal{H}}
\newcommand{\Tcal}{\mathcal{T}}

\newcommand{\Fcal}{\mathcal{F}}
\newcommand{\Pcal}{\mathcal{P}}
\newcommand{\Xcal}{\mathcal{X}}
\newcommand{\Ycal}{\mathcal{Y}}
\newcommand{\Gcal}{\mathcal{G}}

\newcommand{\Mcal}{\mathcal{M}}
\newcommand{\Vcal}{\mathcal{V}}

\newcommand{\Eb}{\mathbf{E}}

\DeclareMathOperator*{\tr}{\mathrm{tr}}

\DeclareMathOperator*{\maxi}{\mathrm{maximize}}

\newcommand{\var}{{\mathrm{Var}}}

\newcommand{\half}{\frac{1}{2}}

\newcommand{\intset}[1]{\cbr{1..n}}

\newtheorem{problem}{Problem}

\newcommand{\mmd}{\mathrm{MMD}}

\newcommand{\Bcal}{\mathcal{B}}

\newcommand{\bdatax}{X}

\begin{document}

\title{A Kernel Method for the Two-Sample Problem}

\author{\name Arthur Gretton \email arthur@tuebingen.mpg.de\\
\addr MPI for Biological Cybernetics\\
Spemannstrasse 38\\
72076, T\"ubingen, Germany
\AND
\name Karsten M. Borgwardt\thanks{This work was carried out while
  K.M.B. was with the Ludwig-Maximilians-Universit\"at M\"unchen.} \email kmb51@cam.ac.uk\\ 
\addr University of Cambridge \\
Department of Engineering \\
Trumpington Street, CB2 1PZ Cambridge, United Kingdom
\AND
\name Malte J. Rasch \email malte@igi.tu-graz.ac.at \\
\addr Graz University of Technology \\
Inffeldgasse 16b/I
8010 Graz, Austria 
\AND
 \name Bernhard~Sch\"olkopf \email  bernhard.schoelkopf@tuebingen.mpg.de\\
\addr MPI for Biological Cybernetics\\
Spemannstrasse 38\\
72076, T\"ubingen, Germany
\AND
 \name Alexander Smola \email alex.smola@gmail.com\\
\addr National ICT Australia\\
Canberra, ACT 0200, Australia
}

\editor{TBA}

\maketitle

\begin{abstract}%
\noindent
We propose a framework for analyzing and comparing distributions, allowing
   us to design statistical tests to determine if two samples are
  drawn from different distributions.  Our test statistic is the
  largest difference in expectations over functions in the unit ball of a
  reproducing kernel Hilbert space (RKHS).  We present two tests based on
  large deviation bounds for the test statistic, while a third is
  based on the asymptotic distribution of this statistic.  The test
  statistic can be computed in quadratic time, although efficient linear time
  approximations are available. 
  Several classical metrics on distributions are recovered when the function
  space used to compute the difference in expectations is allowed to be more general (eg.~a Banach space).
  We apply our two-sample tests  to a variety of
  problems, including attribute matching for databases using the
  Hungarian marriage method, where they perform strongly.  
  Excellent performance is also obtained when comparing distributions over
  graphs, for which these are the first such tests.
\end{abstract}

\medskip


\section{Introduction}

We address the problem of comparing samples from two
probability distributions, by proposing  statistical tests of the
hypothesis that these distributions are different (this is  called 
the two-sample or homogeneity problem). Such tests have application in a
 variety of areas. In bioinformatics, it is of interest to compare
microarray data from identical tissue types as measured by different laboratories,
to detect whether the data may be analysed jointly, or whether differences
in experimental procedure have caused systematic differences in the data distributions.
Equally of interest are comparisons between microarray data from
different tissue types, either to determine whether two subtypes of
cancer may be treated as statistically indistinguishable from a
diagnosis perspective, or to detect differences in healthy and cancerous
tissue.
 In database attribute
matching, it is desirable to merge databases containing multiple
fields, where it is not known in advance which fields correspond: the
fields are matched by maximising the similarity in the distributions
of their entries.

We test whether distributions $p$ and $q$ are different on the basis of
samples drawn from each of them, by finding a well behaved (e.g.\
smooth) function which is large on the points drawn from $p$, and small
(as negative as possible) on the points from $q$. We use as our test
statistic the difference between the mean function values on the two
samples; when this is large, the samples are likely from different
distributions. We call this statistic the Maximum Mean Discrepancy
(MMD).

Clearly the quality of the MMD as a statistic  depends
 on the class $\Fcal$ of smooth functions that define
it.  On one hand, $\Fcal$ must be ``rich enough'' so that the
population MMD vanishes if and only if $p = q$. On the other hand, for
the test to be consistent, $\Fcal$ needs to be ``restrictive''
enough for the empirical estimate of MMD to converge quickly to its
expectation as the sample size increases.  We shall use the unit balls
in universal reproducing kernel Hilbert spaces \citep{Steinwart01b} as
our function classes, since these will be shown to satisfy both of the
foregoing properties (we also review classical metrics on distributions,
namely the Kolmogorov-Smirnov and Earth-Mover's distances,
which are based on different function classes).
On a more practical note, the MMD has a reasonable computational cost, when
compared with other two-sample tests: given $m$ points sampled from $p$
and $n$ from $q$, the cost is $O(m+n)^2$ time. We also propose a less
statistically efficient algorithm  with a computational cost of
$O(m+n)$, which can yield superior performance at a given computational cost
by looking at a larger volume of data.

We define three non-parametric statistical tests based on the MMD.  The first two,
which use distribution-independent uniform convergence bounds, provide
finite sample guarantees of test performance, at the expense of being
conservative in detecting differences between $p$ and $q$. The third
test is based on the asymptotic distribution of the MMD, and is in practice
more sensitive to differences in distribution at small sample sizes. The
present work synthesizes and expands on  results of
\citet{GreBorRasSchSmo07,GreBorRasSchetal07}, \citet{SmoGreSonSch07b},
and \citet{SonZhaSmoGreetal08}\footnote{In particular, most of the proofs
  here were not provided by \citet{GreBorRasSchSmo07}} who in turn build
on the earlier work of \citet{BorGreRasKrietal06}. Note that the
latter addresses only the third kind of test, and that the 
approach of \citet{GreBorRasSchSmo07,GreBorRasSchetal07} employs a more accurate approximation to the asymptotic
distribution of the test statistic.

We begin our presentation in Section \ref{sec:basicStuffAndReview} with
a formal definition of the MMD, and a 
proof that the population MMD is zero if and only if $p=q$ when
$\Fcal$ is the unit ball of a universal RKHS.  
We also review alternative function classes for which the MMD defines a
metric on probability distributions.
In Section \ref{sec:prevWork}, we give an overview of hypothesis testing as it applies to the two-sample problem,
and review other approaches to this problem.
We present our first two hypothesis tests in Section \ref{sec:firstBound}, 
based on two different bounds on the deviation between the population and empirical
$\mmd$.
 We take a
different approach in Section \ref{sec:asymptoticTest}, where we use the
asymptotic distribution of the empirical $\mmd$ estimate as the basis
for a third test. 
When large volumes of data are available, the cost of computing the MMD (quadratic in the sample size)
may  be excessive: we therefore propose in Section \ref{sec:linearTimeStatistic} a modified version of the MMD statistic that has a linear cost in the number of samples, and an associated asymptotic test.
In Section \ref{sec:relatedMethods}, we provide an overview of
 methods related to the MMD in the statistics and machine learning literature.
Finally, in Section \ref{sec:experiments}, we
demonstrate the performance of MMD-based two-sample tests on problems from neuroscience,
bioinformatics, and attribute matching using the Hungarian marriage
method. Our approach performs well on high dimensional data with low sample size;
 in addition, we are able to successfully distinguish distributions on graph data, 
for which ours is the first proposed test.

\section{The Maximum Mean Discrepancy}\label{sec:basicStuffAndReview}

In this section, we present the maximum mean discrepancy (MMD), and describe conditions
under which it is a metric on the space of probability distributions. The MMD is defined in terms
of particular function spaces that witness the difference in distributions: we therefore begin in
 Section \ref{sec:MMDintro} by introducing the MMD for some arbitrary function space. In Section
\ref{sec:mmdInRKHS}, we compute both the population  MMD and two empirical estimates when the associated function space is a 
reproducing kernel Hilbert space, and we derive the RKHS function that witnesses the MMD for a given
pair of distributions in Section \ref{sec:MMDwitness}. Finally, we describe the MMD for more general function
classes in Section \ref{sec:MMDotherFuncClasses}.

\subsection{Definition of the Maximum Mean Discrepancy}\label{sec:MMDintro}
Our goal is to formulate a statistical test that answers the following question: 
\begin{problem}
  \label{prob:problem}
  Let $p$ and $q$ be Borel probability measures defined on a domain $\Xcal$.  Given
  observations $X := \cbr{x_1, \ldots, x_m}$ and $Y := \cbr{y_1, \ldots,
    y_n}$, drawn independently and identically distributed (i.i.d.) from
  $p$ and $q$, respectively, can we decide whether $p \neq q$?
\end{problem}
To start with, we wish to determine a criterion that, in the
population setting, takes on a unique and distinctive value only when
$p=q$.  It will be defined based on Lemma 9.3.2 of \citet{Dudley02}.
\begin{lemma}\label{lem:dudley}
  Let $(\Xcal,d)$ be a metric space, and let $p,q$ be two Borel
  probability measures defined on $\Xcal$. 
  Then $p=q$ if and only if $\Eb_{x\sim p}(f(x))=\Eb_{y \sim q}(f(y))$ for all $f \in C(\Xcal)$, where 
$C(\Xcal)$ is the space of bounded  continuous
functions on $\Xcal$.
\end{lemma}
%
Although $C(\Xcal)$ in principle allows us to identify $p=q$
uniquely, it is not practical to work with such a rich function class in
the finite sample setting.  We thus define a more general class of
statistic, for as yet unspecified function classes $\Fcal$, to measure
the disparity between $p$ and $q$ \citep{ForMou53,Mueller97}.
\begin{definition}
  \label{def:mmd}
  Let $\Fcal$ be a class of functions $f: \Xcal \to \RR$ and let $p,q,X,
  Y$ be defined as above. We define the
  maximum mean discrepancy (MMD)  as
  \begin{equation}
    \label{eq:mmd-a}
    \mmd\sbr{\Fcal, p, q}  := \sup_{f \in \Fcal} 
  \left(\Eb_{x \sim p}[f(x)] -
      \Eb_{y \sim q}[f(y)] \right).
   \end{equation}
\citet{Mueller97} calls this an integral probability metric. A biased  empirical estimate of the MMD is
\begin{equation}
    \label{eq:mmd-e}
    \mmd_b\sbr{\Fcal, X, Y}  := \sup_{f \in \Fcal} 
\left(      \frac{1}{m} \sum_{i=1}^m f(x_i) -
      \frac{1}{n} \sum_{i=1}^n f(y_i)  \right).
  \end{equation}
\end{definition}
The empirical MMD defined above has an upward bias (we will
define an unbiased statistic in the following section).
We must now identify a function class that is rich enough to uniquely
identify whether $p=q$, yet restrictive enough to provide useful finite
sample estimates (the latter property will be established in subsequent
sections).  

\subsection{The MMD in Reproducing Kernel Hilbert Spaces\label{sec:mmdInRKHS}}

If $\Fcal$ is the unit ball in a reproducing kernel Hilbert space
$\Hcal$, the empirical MMD can be computed very
efficiently. This will be the main approach we pursue in the present study. 
Other possible function classes $\Fcal$ are discussed at the end of this section. 
We will refer to $\Hcal$ as universal whenever $\Hcal$,
defined on a compact metric space $\Xcal$ and with associated kernel $k:
\Xcal^2 \to \RR$, is dense in $C(\Xcal)$ with respect to the $L_\infty$
norm. It is shown in \cite{Steinwart01b} that Gaussian and Laplace
kernels are universal. We have the following result:
\begin{theorem}
  \label{th:stronger}
  Let $\Fcal$ be a unit ball in a universal RKHS $\Hcal$, defined on the
  compact metric space $\Xcal$, with associated kernel
  $k(\cdot,\cdot)$. Then $\mmd\sbr{\Fcal, p,q} = 0$ if and only if $p =
  q$.
\end{theorem}
\begin{proof}
  It is clear that $\mmd\sbr{\Fcal,p,q}$ is zero if $p=q$. We prove the
  converse by showing that $\mmd\sbr{C(\Xcal),p,q}=D$ for some $D>0$
  implies $\mmd\sbr{\Fcal,p,q} > 0$: this is equivalent to
  $\mmd\sbr{\Fcal,p,q}=0$ implying $\mmd\sbr{C(\Xcal),p,q}=0$ (where
  this last result implies $p=q$ by Lemma \ref{lem:dudley}, noting that
  compactness of the metric space $\Xcal$ implies its separability). 
Let $\Hcal$ be the universal RKHS of which $\Fcal$ is the unit ball.
If $\mmd\sbr{C(\Xcal),p,q}=D$,
then there exists some $\tilde{f}\in C(\Xcal)$ for which $\Eb_{p}\sbr{\tilde{f}}-\Eb_{q}\sbr{\tilde{f}}\ge D/2$.
We know that $\Hcal$ is dense in $C(\Xcal)$ with respect
to the $L_{\infty}$ norm: this means that for $\epsilon = D/8$,
we can find some $f^{*}\in\Hcal$ satisfying $\left\Vert f^{*}-\tilde{f}\right\Vert _{\infty}<\epsilon$.
Thus, we obtain $\abr{\Eb_p \sbr{f^*} - \Eb_p\sbr{\tilde{f}}} <
\epsilon$ and consequently  
$$
  \abr{\Eb_{p}\sbr{f^{*}}-\Eb_{q}\sbr{f^{*}}}  >  
  \abr{\Eb_{p}\sbr{\tilde{f}}-\Eb_{q}\sbr{\tilde{f}}} - 2\epsilon
> \textstyle \frac{D}{2} - 2 \frac{D}{8} = \frac{D}{4} > 0.
$$
Finally, using $\left\Vert f^{*}\right\Vert _{\Hcal}<\infty$,
we have
\begin{align*}
  \sbr{\Eb_{p}\sbr{f^{*}}-\Eb_{q}\sbr{f^{*}}}/{\nbr{f^{*}}_{\Hcal}} 
  \ge D / (4 \nbr{f^{*}}_{\Hcal})>0,
\end{align*}
and hence $\mmd\sbr{\Fcal,p,q}>0$.
\end{proof}
We now review some properties of $\Hcal$ that will allow us to
express the MMD in a more easily computable form \citep{SchSmo02}. 
Since $\Hcal$ is an RKHS, the operator of evaluation $\delta_x$ mapping
$f\in \Hcal$ to $f(x)\in\RR$ is continuous. Thus, by the Riesz representation
theorem, there is a feature mapping $\phi(x)$ from $\Xcal$ to $\RR$ such
that $f(x)=\inner{f}{\phi(x)}_{\Hcal}$. Moreover, $\inner{\phi(x)}{\phi(y)}_{\Hcal} = k(x,y)$,
where $k(x,y)$ is a positive definite kernel function. 
The following lemma is due to \citet{BorGreRasKrietal06}.
\begin{lemma}\label{le:simpleMMD}
Denote the expectation of $\phi(x)$ by $\mu_p := \Eb_{p}
  \sbr{\phi(x)}$ (assuming its existence).\footnote{A sufficient condition
    for this is $\|\mu_p\|^2_\Hcal<\infty$, which is rearranged as
    $\Eb_{p}[k(x,x')]<\infty$, where $x$ and $x'$ are independent random
    variables drawn according to $p$. In other words, $k$ is a trace
    class operator with respect to the measure $p$.}  
Then \vspace{-5mm}
\begin{align}
  \label{eq:norm}
  \mmd[\Fcal,p,q] & = \sup_{\nbr{f}_\Hcal \leq 1} 
  \inner{\mu[p] - \mu[q]}{f} 
  = \nbr{ \mu[p] - \mu[q] }_{\Hcal}.
\end{align}
\end{lemma}
\begin{proof}
\begin{eqnarray*}
   \mmd^2[\Fcal,p,q] 
  & = & \left[\sup_{\nbr{f}_\Hcal \leq 1} 
\left(  \Eb_p\sbr{f(x)} - \Eb_q\sbr{f(y)} \right) \right]^2\\
  & = & \left[\sup_{\nbr{f}_\Hcal \leq 1} 
\left(  \Eb_p\sbr{\inner{\phi(x)}{f}_{\Hcal}} - \Eb_q\sbr{\inner{\phi(y)}{f}_{\Hcal}} \right) \right]^2\\
  & = & \left[\sup_{\nbr{f}_\Hcal \leq 1} 
  \inner{\mu_p - \mu_q }{f}_{\Hcal} \right]^2
  = \nbr{ \mu_p - \mu_q }^2_{\Hcal} 
\end{eqnarray*}
\end{proof}
Given we are in an RKHS, the norm $\nbr{ \mu_p - \mu_q }^2_{\Hcal}$
may easily be computed in terms of kernel functions. This leads to a first  empirical
estimate of the MMD, which is unbiased.
\begin{lemma}
  \label{lem:rkhs-mmd}
Given $x$ and $x'$ independent random variables with distribution $p$, 
 and $y$ and $y'$ independent random variables with distribution $q$,
the population $\mmd^2$ is
\begin{equation}
\label{eq:mmd-a-rkhs}
\mmd^2\sbr{\Fcal, p, q}  = \Eb_{x,x' \sim p} \sbr{k(x,x')} 
      -2\Eb_{x \sim p, y \sim q} \sbr{k(x,y)}
      + \Eb_{y,y' \sim q} \sbr{k(y,y')}.
\end{equation}
Let  $Z:=({z}_{1},\ldots,{z}_{m})$ be $m$ i.i.d. random variables, where $z_i:=(x_i,y_i)$ (i.e. we assume $m=n$). An \emph{unbiased} empirical estimate of $\mmd^2$ is
\begin{equation}
    \label{eq:mmd-e-rkhs-unbiased}
\mmd_u^2 \sbr{\Fcal,X,Y}   =  \frac{1}{(m)(m-1)}\sum_{i\neq j}^{m} 
h({z}_i,{z}_{j}),
\end{equation}
which is a one-sample U-statistic with $h(z_i,z_j):=k(x_i, x_j) + k(y_i, y_j) - k(x_i, y_j) - k(x_j, y_i)$
(we define $h(z_i,z_j)$ to be symmetric in its arguments due to requirements that will arise in  Section \ref{sec:asymptoticTest}).
\end{lemma}
\begin{proof}
Starting from the expression for $\mmd^2[\Fcal,p,q]$ in Lemma \ref{le:simpleMMD},
\begin{eqnarray*}
   \mmd^2[\Fcal,p,q]  
  &=& \nbr{ \mu_p - \mu_q }^2_{\Hcal} \\
  &=& \inner{\mu_p}{\mu_p}_\Hcal + \inner{\mu_q}{\mu_q}_\Hcal - 2\inner{\mu_p}{\mu_q}_\Hcal \\
  &=& \Eb_p \inner{\phi(x)}{\phi(x')}_\Hcal + \Eb_q \inner{\phi(y)}{\phi(y')}_\Hcal 
  - 2\Eb_{p,q}\inner{\phi(x)}{\phi(y)}_\Hcal,
\end{eqnarray*}
The proof is completed by applying
$\inner{\phi(x)}{\phi(x')}_\Hcal = k(x,x')$; the empirical estimate
follows straightforwardly.
\end{proof}
The empirical statistic is an unbiased estimate of $\mmd^2$, although it
does not have minimum variance, since we are ignoring the cross-terms
$k(x_i, y_i)$ of which there are only $O(n)$. The minimum variance
estimate is almost identical, though \citep[Section
5.1.4]{Serfling80}.

The biased statistic in (\ref{eq:mmd-e}) may also be easily computed following the above reasoning.
Substituting the empirical estimates  $\mu[X] :=
\frac{1}{m} \sum_{i=1}^m \phi(x_i)$ and $\mu[Y]:=\frac{1}{n} \sum_{i=1}^n \phi(y_i)$ of the feature space means based on respective samples $X$ and $Y$,  we obtain
\begin{equation}
 \label{eq:mmd-e-rkhs}
    \mmd_b\sbr{\Fcal, X, Y}  = \sbr{\frac{1}{m^2} \sum_{i,j=1}^m {k(x_i,x_j)} 
      -\frac{2}{m n} \sum_{i,j=1}^{m,n} k(x_i, y_j)
   +  \frac{1}{n^2} \sum_{i,j=1}^n {k(y_i,y_j)}}^{\frac{1}{2}}.
\end{equation}

 Intuitively we expect the empirical test statistic $\mmd[\Fcal, X, Y]$, 
whether biased or unbiased, to be small if $p = q$,
and  large if the distributions are far apart. 
It costs $O((m+n)^2)$ time to compute both statistics.

Finally, we note that \citet{HarBacMou08} recently proposed a modification 
of the kernel MMD statistic in Lemma \ref{le:simpleMMD},
by scaling the feature space mean distance using the inverse within-sample covariance operator, 
thus employing the kernel Fisher discriminant as a statistic for testing homogeneity. This
statistic is shown to be related to the $\chi^2$ divergence.

\subsection{Witness Function of the MMD for RKHSs\label{sec:MMDwitness}}

\begin{figure}
\begin{center}
\includegraphics[width=0.5\textwidth]{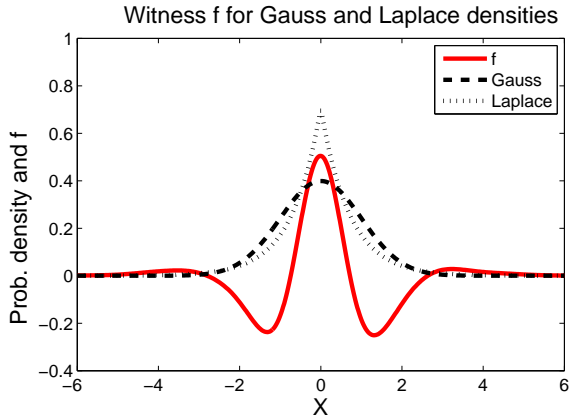}
\end{center}
\caption{Illustration of the function maximizing the mean discrepancy in the case
  where a Gaussian is being compared with a Laplace distribution. Both
  distributions have zero mean and unit variance.
The function $f$ that witnesses the MMD has been
  scaled for plotting purposes, and was computed empirically on the
  basis of $2\times 10^4$ samples, using a Gaussian kernel with
  $\sigma=0.5$.\label{fig:mmdDemo1D}}
\end{figure}

It is also instructive to consider the witness $f$ which is  chosen by
MMD to exhibit the maximum discrepancy between the two distributions. 
The population $f$ and its empirical estimate $\hat{f}(x)$ are respectively
\begin{equation*}
\begin{array}{cclcl}
 f(x)  &\propto 
  &\inner{\phi(x)}{\mu[p] - \mu[q]} 
  &= 
  &\Eb_{x' \sim p} \sbr{k(x,x')} -   \Eb_{x' \sim q} \sbr{k(x,x')}  \\
\hat{f}(x) &\propto 
&\inner{\phi(x)}{\mu[X] - \mu[Y]}
&= 
&\frac{1}{m} \sum_{i=1}^m k(x_i, x) - 
  \frac{1}{n} \sum_{i=1}^n k(y_i, x).   
\end{array}
\end{equation*}
This follows from the fact that the unit vector $v$ maximizing
$\inner{v}{x}_{\Hcal}$ in a Hilbert space is $v = x/\nbr{x}$. 

We illustrate the behavior of MMD in Figure \ref{fig:mmdDemo1D} using a
one-dimensional example. The data $X$ and $Y$ were generated from
distributions $p$ and $q$ with equal means and variances, with $p$ 
Gaussian and $q$  Laplacian. We chose
$\Fcal$ to be the unit ball in an RKHS using the Gaussian kernel. We observe that the
function $f$ that witnesses the MMD --- in other words, the function
maximizing the mean discrepancy in (\ref{eq:mmd-a}) --- is smooth, positive
where the Laplace density exceeds the Gaussian density (at the center
and tails), and negative where the Gaussian density is larger. Moreover,
the magnitude of $f$ is a direct reflection of the amount by which one
density exceeds the other, insofar as the smoothness constraint permits
it.

\subsection{The MMD in Other Function Classes\label{sec:MMDotherFuncClasses}}

The definition of the maximum mean discrepancy is by no means limited to
RKHS. In fact, any function class $\Fcal$ that comes with uniform
convergence guarantees and  is sufficiently powerful will enjoy the
above properties. 
\begin{definition}
  Let $\Fcal$ be a subset of some vector space. 
  The star $S[\Fcal]$ of a set $\Fcal$ is 
  \begin{align*}
    S[\Fcal] := \cbr{\alpha x | x \in \Fcal \text{ and } \alpha \in [0, \infty)}
  \end{align*}
\end{definition}
\begin{theorem}
  \label{th:stronger-star}
  Denote by $\Fcal$ the subset of some vector space of functions from
  $\Xcal$ to $\RR$ for which $S[\Fcal] \cap C(\Xcal)$ is dense in $C(\Xcal)$ with
  respect to the $L_\infty(\Xcal)$ norm. Then $\mmd\sbr{\Fcal, p,q} = 0$
  if and only if $p = q$.

  Moreover, under the above conditions $\mmd[\Fcal, p, q]$ is a
  metric on the space of probability distributions. Whenever the star of
  $\Fcal$ is \emph{not} dense, $\mmd$ is a pseudo-metric space.\footnote{According to \citet[][p. 26]{Dudley02} a metric $d(x,y)$ satisfies the following four properties: symmetry, triangle inequality, $d(x,x)=0$, and $d(x,y)=0\implies x=y$. A pseudo-metric only satisfies the first three properties.} 
\end{theorem}
\begin{proof}
  The first part of the proof is almost identical to that of
  Theorem~\ref{th:stronger} and is therefore omitted. To see the second
  part, we only need to prove the triangle inequality. We have
  \begin{align*}
    \sup_{f \in \Fcal} \abr{E_p f - E_q f} + 
    \sup_{g \in \Fcal} \abr{E_q g - E_r g} &\geq   
    \sup_{f \in \Fcal} \sbr{\abr{E_p f - E_q f} + 
      \abr{E_q f - E_r }} \\
 &\geq \sup_{f \in \Fcal} 
    \abr{E_p f - E_r f}.
  \end{align*}
  The first part of the theorem establishes that $\mmd[\Fcal,
  p, q]$ is a metric, since only for $p = q$ do we have $\mmd[\Fcal, p, q]
  = 0$. 
\end{proof}
%
Note that any uniform convergence statements in terms of $\Fcal$
allow us immediately to characterize an estimator of $\mmd(\Fcal, p, q)$
explicitly. The following result shows how (we will refine this
reasoning for the RKHS case in Section~\ref{sec:firstBound}). 
\begin{theorem}
  \label{th:general}
  Let $\delta \in (0, 1)$ be a confidence level and assume that for some
  $\epsilon(\delta, m, \Fcal)$ the following holds for samples
  $\cbr{x_1, \ldots, x_m}$ drawn from $p$:
  \begin{align}
    \Pr\cbr{\sup_{f \in \Fcal} \abr{\Eb_p[f] - \frac{1}{m} \sum_{i=1}^m
        f(x_i)} > \epsilon(\delta, m, \Fcal)} \leq \delta.
  \end{align}
  In this case we have that 
  \begin{align}
    \Pr\cbr{ \abr{\mmd[\Fcal,p,q] - \mmd_b[\Fcal,X,Y]} > 2
      \epsilon(\delta/2,m,\Fcal)} \leq \delta.
  \end{align}
\end{theorem}
\begin{proof}
  The proof works simply by using convexity and suprema as follows:
  \begin{align*}
    & \abr{\mmd[\Fcal,p,q] - \mmd_b[\Fcal,X,Y]} \\
    = & \abr{\sup_{f \in \Fcal} \abr{\Eb_p[f] - \Eb_q[f]} - 
      \sup_{f \in \Fcal} \abr{\frac{1}{m} \sum_{i=1}^m f(x_i) - 
        \frac{1}{n} \sum_{i=1}^n f(y_i)}} \\
    \leq & \sup_{f \in \Fcal} \abr{\Eb_p[f] - \Eb_q[f] - 
      \frac{1}{m} \sum_{i=1}^m f(x_i) + 
      \frac{1}{n} \sum_{i=1}^n f(y_i)} \\
    \leq & 
    \sup_{f \in \Fcal} \abr{\Eb_p[f] - \frac{1}{m} \sum_{i=1}^m f(x_i)} + 
    \sup_{f \in \Fcal} \abr{\Eb_q[f] - \frac{1}{n} \sum_{i=1}^n f(y_i)}. 
  \end{align*}
  Bounding each of the two terms via a uniform convergence bound
  proves the claim. 
\end{proof}
This shows that $\mmd_b[\Fcal, X, Y]$ can be used to estimate $\mmd[\Fcal,
p, q]$ and that the quantity is asymptotically
unbiased.

\begin{remark}[Reduction to Binary Classification]
  Any classifier which maps a set of observations $\cbr{z_i, l_i}$ with
  $z_i \in \Xcal$ on some domain $\Xcal$ and labels $l_i \in \cbr{\pm
    1}$, for which uniform convergence bounds  exist on the convergence of the
  empirical loss to the expected loss, can be used to obtain a
  similarity measure on distributions --- simply assign $l_i = 1$ if $z_i \in X$ and $l_i =
  -1$ for $z_i \in Y$ and find a classifier which is able to separate
  the two sets. In this case maximization of $\Eb_p[f] - \Eb_q[f]$ is
  achieved by ensuring that as many $z \sim p(z)$ as possible correspond
  to $f(z) = 1$, whereas for as many $z \sim q(z)$ as possible we have
  $f(z) = -1$. Consequently neural networks, decision trees, boosted
  classifiers and other objects for which uniform convergence bounds can
  be obtained can be used for the purpose of distribution comparison. 
 For instance, \citet[][Section 4]{DavBliCraPer07} use the error of a hyperplane classifier 
to approximate the $\mathcal{A}$-distance
between distributions of \citet{KifDavGeh04}.
\end{remark}

\subsection{Examples of Non-RKHS Function Classes\label{sec:nonRKHSfunctionClasses}}

Other function spaces $\Fcal$ inspired by the statistics literature 
can also be considered in defining the MMD. Indeed, Lemma \ref{lem:dudley}
defines an MMD with $\Fcal$ the space of bounded continuous real-valued functions, which
is a Banach space with the supremum norm \citep[][p. 158]{Dudley02}.
 We now describe two further metrics on the space of probability distributions, 
  the Kolmogorov-Smirnov and Earth Mover's distances, and their associated function classes.

\subsubsection{Kolmogorov-Smirnov Statistic}

The Kolmogorov-Smirnov (K-S) test is probably one of the most famous
two-sample tests in statistics. It works for random variables $x \in
\RR$ (or any other set for which we can establish a total
order). Denote by $F_p(x)$ the cumulative distribution function of $p$
and let $F_X(x)$ be its empirical counterpart, that is
\begin{align*}
  F_p(z) := \Pr\cbr{x \leq z \text{ for } x \sim p(x)} 
  \text{ and }
  F_X(z) := \frac{1}{|X|} \sum_{i=1}^m 1_{z \leq x_i}.
\end{align*}
It is clear that $F_p$ captures the properties of $p$. The Kolmogorov metric
is simply the $L_\infty$ distance $\nbr{F_X - F_Y}_\infty$ for two sets
of observations $X$ and $Y$. \cite{Smirnov39} showed that for $p = q$
the limiting distribution of the empirical cumulative distribution
functions satisfies
\begin{align}
  \label{eq:ks-asympt}
  \lim_{m,n \to \infty} \Pr\cbr{\sbr{\textstyle
      \frac{mn}{m+n}}^{\frac{1}{2}}
    \nbr{F_X - F_Y}_\infty > x
  } = 2 \sum_{j=1}^\infty (-1)^{j-1} e^{-2j^2 x^2} \text{ for } x \geq 0.
\end{align}
This allows for an efficient characterization of the distribution under
the null hypothesis $\Hcal_0$. Efficient numerical approximations to
\eq{eq:ks-asympt} can be found in numerical analysis handbooks
\citep{PreTeuVetFla94}. The distribution under the alternative, $p \neq
q$, however, is unknown.

The Kolmogorov metric is, in fact, a special
instance of $\mmd[\Fcal, p, q]$ for a certain Banach space \citep[][Theorem 5.2]{Mueller97}
\begin{proposition}
  \label{th:ks}
  Let $\Fcal$ be the class of functions $\Xcal \to \RR$ of bounded
  variation\footnote{A  function $f$ defined on $[a,b]$
 is of bounded variation $C$ if the total variation is bounded by $C$,
i.e.  the supremum over all sums
$$
\sum_{1\le i \le n} |f(x_i) - f(x_{i-1})|,
$$
where $a\le x_0 \le \ldots \le x_n \le b$ \citep[][p. 184]{Dudley02}.
}
 1. Then 
  $\mmd[\Fcal, p, q] = \nbr{F_p - F_q}_\infty$.
\end{proposition}

\subsubsection{Earth-Mover Distances}

Another class of distance measures on distributions that may be written
as an MMD are the Earth-Mover distances. We assume $(\mathcal{X},d)$
is a separable metric space, and define $\mathcal{P}_{1}(\mathcal{X})$
to be the space of probability measures on $\mathcal{X}$ for which
$\int d(x,z)dp(z)<\infty$ for all $p\in\mathcal{P}_{1}(\mathcal{X})$
and $x\in\mathcal{X}$ (these are the probability measures for which
$\Eb\left|x\right|<\infty$ when $\mathcal{X}=\RR$). We then have
the following definition \citep[][p. 420]{Dudley02}.
\begin{definition}[Monge-Wasserstein metric]
Let $p\in\mathcal{P}_{1}(\mathcal{X})$ and $q\in\mathcal{P}_{1}(\mathcal{X})$.
The Monge-Wasserstein distance is defined as\[
W(p,q):=\inf_{\mu\in M(p,q)}\int d(x,y)d\mu(x,y),\]
where $M(p,q)$ is the set of joint distributions on $\mathcal{X}\times\mathcal{X}$
with marginals $p$ and $q$. 
\end{definition}
We may interpret this as the cost (as represented by the metric $d(x,y)$)
of transferring mass distributed according to $p$ to a distribution
in accordance with $q$, where $\mu$ is the movement schedule. 
In general, a large
variety of costs of moving mass from $x$ to $y$ can be used, such as
psychooptical similarity measures in image retrieval
\citep{RubTomGui00}.
The
following theorem holds \citep[][Theorem 11.8.2]{Dudley02}.
\begin{theorem}[Kantorovich-Rubinstein]\label{thm:KantorovichRubinstein}
 Let $p\in\mathcal{P}_{1}(\mathcal{X})$ and
$q\in\mathcal{P}_{1}(\mathcal{X})$, where $\Xcal$ is separable. Then a metric on $\mathcal{P}_{1}(S)$
is defined as\[
W(p,q)=\left\Vert p-q\right\Vert _{L}^{*}=\sup_{\left\Vert f\right\Vert _{L}\le1}\left|\int f\, d(p-q)\right|,\]
where\[
\left\Vert f\right\Vert _{L}:=\sup_{x\neq y\,\in\, \Xcal}\frac{\left|f(x)-f(y)\right|}{d(x,y)}\]
is the Lipschitz seminorm%
\footnote{A seminorm satisfies the requirements of a norm besides $\left\Vert x\right\Vert =0$
only for $x=0$ \citep[][p. 156]{Dudley02}.}
for real valued $f$ on $\Xcal$.
\end{theorem}
%
A simple
example of this theorem is as follows \citep[][Exercise 1,
p. 425]{Dudley02}.
%
\begin{example}
Let $\mathcal{X}=\RR$ with associated $d(x,y)=\left|x-y\right|$.
Then given $f$ such that $\left\Vert f\right\Vert _{L}\le1$, we
use integration by parts to obtain\[
\left|\int f\, d(p-q)\right|=\left|\int(F_{p}-F_{q})(x)f'(x)dx\right|\le\int\left|(F_{p}-F_{q})\right|(x)dx,\]
where the maximum is attained for the function $g$ with derivative
$g'=2\, 1_{F_{p}>F_{q}}-1$ (and for which $\left\Vert g\right\Vert _{L}=1$).
We recover the $L_{1}$ distance between distribution functions,\[
W(P,Q)=\int\left|(F_{p}-F_{q})\right|(x)dx.\]
\end{example}
One may further generalize Theorem \ref{thm:KantorovichRubinstein} to the set of all laws $\mathcal{P}(\mathcal{X})$
on arbitrary metric spaces $\mathcal{X}$ \citep[][Proposition 11.3.2]{Dudley02}.
\begin{definition}[Bounded Lipschitz metric]
 Let $p$ and $q$ be laws on a metric space
$\mathcal{X}$. Then\[
\beta(p,q):=\sup_{\left\Vert f\right\Vert _{BL}\le1}\left|\int f\, d(p-q)\right|\]
is a metric on $\mathcal{P}(\mathcal{X})$, where $f$ belongs to
the space of bounded Lipschitz functions with norm\[
\left\Vert f\right\Vert _{BL}:=\left\Vert f\right\Vert _{L}+\left\Vert f\right\Vert _{\infty}.\]
\end{definition}

\section{Background Material}\label{sec:prevWork}

We now present three background results. First, we introduce the terminology
used in statistical hypothesis testing. Second, we demonstrate via an example that
even for tests which have asymptotically no error, one cannot guarantee performance
at any fixed sample size without making assumptions about the distributions.
Finally, we briefly review some earlier approaches to the two-sample problem.

\subsection{Statistical Hypothesis Testing}
Having described a metric on probability distributions (the MMD) based 
on distances between their Hilbert space embeddings, and empirical
estimates (biased and unbiased) of this metric, we now address
the problem of determining whether the empirical MMD shows 
a \emph{statistically significant} difference between distributions.
To this end, we briefly describe the framework of
statistical hypothesis testing as it applies in the present context,
following \citet[Chapter 8]{CasBer02}. Given i.i.d. samples $X\sim p$ of size
$m$ and $Y\sim q$ of size $n$, the statistical test,
$\Tcal(X,Y)\,:\,\Xcal^m \times \Xcal^n\mapsto\{0,1\}$ is used to distinguish between
the null hypothesis $\Hcal_0\,:\,p=q$ and the alternative hypothesis
$\Hcal_1\,:\,p\neq q$.  This is achieved by comparing the test statistic\footnote{This may be biased or unbiased.}
$\mmd[\Fcal,X,Y]$ with a particular threshold: if the threshold is
exceeded, then the test rejects the null hypothesis (bearing in mind
that a zero population MMD indicates $p=q$).
The acceptance region of the test is thus defined as the set of real numbers below the threshold.
 Since the test is based on
finite samples, it is possible that an incorrect answer will be
returned: we define the Type I error as the probability of rejecting $p=q$  based on the observed sample, despite the null hypothesis having generated the data. Conversely, the Type II error is the probability of accepting $p=q$ despite the underlying distributions being different.
The
\emph{level} $\alpha$ of a test is an upper bound on the Type I error: this is
a design parameter of the test, and is used to set the threshold to
which we compare the test statistic (finding the test threshold for a
given $\alpha$ is the topic of Sections \ref{sec:firstBound} and
\ref{sec:asymptoticTest}). A consistent test achieves a level $\alpha$,
and a Type II error of zero, in the large sample limit. We will see that
 the tests proposed in this paper are consistent.

\subsection{A Negative Result}

Even if a test is consistent, it is not possible to distinguish distributions
with high probability at a given, fixed sample size (i.e., to provide guarantees on the Type II error),
without prior assumptions as to the nature of the difference between $p$ and $q$.
This is true \emph{regardless} of the two-sample test used.
There are several ways to illustrate this, which each give different insight into the kinds of differences that might be undetectable for a given number of samples.
  The following
example\footnote{This is a variation of a construction for independence
  tests, which was suggested in a private communication by John
  Langford.} is one such illustration.

\begin{example}
  Assume that we have a distribution $p$ from which we draw $m$ iid
  observations. Moreover, we construct a distribution $q$ by drawing
  $m^2$ iid observations from p and subsequently defining a discrete
  distribution over these $m^2$ instances with probability $m^{-2}$
  each. It is easy to check that if we now draw $m$ observations from
  $q$, there is at least a ${{m^2} \choose m} \frac{m!}{m^{2m}} >
  1-e^{-1} > 0.63$ probability that we thereby will have effectively
  obtained an $m$ sample from $p$. 
Hence no test will be able to
  distinguish samples from $p$ and $q$ in this case. We could make the
  probability of detection arbitrarily small by increasing the size of
  the sample from which we construct $q$.
\end{example}

\subsection{Previous Work\label{sec:otherTestsReview}}

We next give a brief overview of some earlier approaches to the two sample
problem for multivariate data.  Since our later experimental comparison
is with respect to certain of these methods, we give abbreviated
algorithm names in italics where appropriate: these should be used as a
key to the tables in Section \ref{sec:experiments}.  A generalisation of
the Wald-Wolfowitz runs test to the multivariate domain was proposed and
analysed by \cite{FriRaf79,HenPen99} \emph{(FR Wolf)}, and involves
counting the number of edges in the minimum spanning tree over the
aggregated data that connect points in $X$ to points in $Y$. The
resulting test relies on the asymptotic normality of the test statistic,
and this quantity is not distribution-free under the null hypothesis for
finite samples (it depends on $p$ and $q$). The computational cost of
this method using Kruskal's algorithm is $O((m+n)^2\log(m+n))$, although
more modern methods improve on the $\log(m+n)$ term. See \cite{Chazelle00}
for details. \citet{FriRaf79} claim that calculating the matrix of
distances, which costs $O((m+n)^2)$, dominates their computing time;
we return to this point in our experiments (Section \ref{sec:experiments}).  Two possible
generalisations of the Kolmogorov-Smirnov test to the multivariate case
were studied in \cite[]{Bickel69,FriRaf79}.  The approach of Friedman and
Rafsky \emph{(FR Smirnov)} in this case again requires a minimal spanning
tree, and has a similar cost to their multivariate runs test.

A more recent multivariate test was introduced by \citet{Rosenbaum05}. This
entails computing the minimum distance non-bipartite matching over the
aggregate data, and using the number of pairs containing a sample from
both $X$ and $Y$ as a test statistic. The resulting statistic is
distribution-free under the null hypothesis at finite sample sizes, in
which respect it is superior to the Friedman-Rafsky test; on the other
hand, it costs $O((m+n)^3)$ to compute.  Another distribution-free
test \emph{(Hall)} was proposed by \citet{HalTaj02}:
for each point from $p$,
it requires computing the closest points in the aggregated data, and counting how many of these are from $q$ (the procedure is repeated for each point from $q$ with respect to points from $p$).
 As we shall see in our experimental
comparisons, the test statistic is costly to compute; \citet{HalTaj02}
consider only tens of points in their experiments.

Yet another approach is to use some distance (e.g. $L_1$ or $L_2$)
between Parzen window estimates of the densities as a test statistic
\citep{AndHalTit94,BiaGyo05}, based on the asymptotic distribution of
this distance given $p=q$. When the $L_2$ norm is used, the test
statistic is related to those we present here, although it is arrived at
from a different perspective.
Briefly, the test of \cite{AndHalTit94} is obtained in a more restricted setting
where the RKHS kernel is an inner product between Parzen windows. Since we are not doing density
estimation, however, we need not decrease the kernel width as the sample grows. In fact, decreasing
the kernel width reduces the convergence rate of the associated two-sample test, compared with
the $(m + n)^{-1/2}$ rate for fixed kernels.
We provide more detail in Section \ref{sec:MMDparzen}.
The $L_1$ approach of
\cite{BiaGyo05} \emph{(Biau)}  requires the space to be  partitioned
 into a grid of bins, which becomes difficult or impossible for high
 dimensional problems. Hence we use this test only for low-dimensional problems
in our experiments.

\section{Tests Based on Uniform Convergence Bounds}\label{sec:firstBound}

In this section, we introduce two statistical tests of independence which have
exact performance guarantees
at finite sample sizes, based on uniform convergence bounds. The first, in Section \ref{sec:boundBiasedStat}, uses the \citet{McDiarmid89}
bound on the biased MMD statistic, and the second, in Section \ref{sec:boundUnbiasedStat}, uses a \cite{Hoeffding63} 
bound for the unbiased statistic. 

\subsection{Bound on the Biased Statistic and Test\label{sec:boundBiasedStat}}

We establish two properties of the MMD, from which we derive a hypothesis test. First, we show
that regardless of whether or not $p=q$, the empirical MMD converges in
probability at rate $O((m+n)^{-\frac{1}{2}})$ to its population value. This
shows the consistency of statistical tests based on the MMD.  Second,
we give probabilistic bounds for large deviations of the empirical MMD
in the case $p=q$.  These bounds lead directly to a threshold for our
first hypothesis test.
%
%
We begin our discussion of the convergence of $\mmd_b[\Fcal, X, Y]$ 
to $\mmd[\Fcal, p,q]$.  
\begin{theorem}\label{th:mmd-diff}
  Let $p,q,X,Y$ be defined as in 
  Problem~\ref{prob:problem}, and assume $0\le k(x,y)\le K$.
  Then 
  \begin{align*}
    \Pr\cbr{\abr{\mmd_b[\Fcal, X,Y] - \mmd[\Fcal, p, q]} > 
2\left( (K/m)^{\frac{1}{2}} + (K/n)^{\frac{1}{2}} \right) + \epsilon}
    \leq 2\exp\rbr{\textstyle \frac{-\epsilon^2 m n}{2 K (m+n)}}. 
  \end{align*}
\end{theorem}
See Appendix \ref{sec:pqdiffproof} for proof.
Our next goal is to refine this result in a way that allows us to
define a test threshold under the null hypothesis $p=q$. Under this
circumstance, the constants in the exponent are slightly improved.
\begin{theorem}\label{th:mmd-same-a}
  Under the conditions of Theorem~\ref{th:mmd-diff} where additionally
  $p = q$ and $m = n$,
  \begin{align}
\nonumber
    \mmd_b[\Fcal, X,Y] \leq  
\underset{B_1(\Fcal,p)}
{\underbrace{  m^{-\half} \sqrt{2 \Eb_p\sbr{k(x,x) - k(x,x')}}}}
 + \epsilon
\leq
 \underset{B_2(\Fcal,p)}
{\underbrace{ (2K/m)^{1/2}}}
 + \epsilon,
  \end{align}
  both with probability at least $1-\exp\rbr{-\frac{\epsilon^2 m}{4 K}}$  (see Appendix \ref{sec:largeDevUnderH0} for the proof).
\end{theorem}
In this theorem, we illustrate two possible
bounds $B_1(\Fcal,p)$ and $B_2(\Fcal,p)$ on the bias in the empirical estimate (\ref{eq:mmd-e-rkhs}).
The first inequality is interesting inasmuch as it provides a link between
the bias bound $B_1(\Fcal,p)$ and kernel size (for instance, if we were to use a Gaussian
kernel with large $\sigma$, then $k(x,x)$ and $k(x,x')$ would likely
be close, and the bias small). In the context of testing, however,
we would need to provide an additional bound to show convergence 
of an empirical estimate of $B_1(\Fcal,p)$ to its population equivalent.
%
%
%
%
%
%
%
%
Thus, in the following test for $p=q$ based on Theorem \ref{th:mmd-same-a},  we use $B_2(\Fcal,p)$ to bound the bias.\footnote{Note that we use a tighter bias bound than \cite{GreBorRasSchSmo07}.}
\begin{corollary}\label{lem:firstTest}
A hypothesis test of level
$\alpha$ for the null hypothesis $p=q$, that is, for $\mmd[\Fcal, p,q]
=0$, has the acceptance region 
$
\mmd_b[\Fcal, X,Y] 
<
 \sqrt{2K/m} \left( 1 + \sqrt{2\log \alpha^{-1}}  \right).
$
\end{corollary}
We emphasise that Theorem \ref{th:mmd-diff} guarantees the consistency
of the test, and that the Type II error probability decreases to zero at
rate $O(m^{-\frac{1}{2}})$, assuming $m=n$.  To put this convergence
rate in perspective, consider a test of whether two normal distributions
have equal means, given they have unknown but equal variance
\cite[Exercise 8.41]{CasBer02}.  In this case, the test statistic has a
Student-$t$ distribution with $n+m-2$ degrees of freedom, and its error
probability converges at the same rate as our test.

It is worth noting that bounds may be obtained for the
deviation between expectations $\mu[p]$ and the empirical means $\mu[X]$
in a completely analogous fashion. The proof requires
symmetrization by means of a \emph{ghost sample}, i.e.\ a second set of
observations drawn from the same distribution. While not the key focus
of the present paper, such bounds can be used in the design of inference
principles based on moment matching
\cite[]{AltSmo06,DudSch06,DudPhiSch04}. 


\subsection{Bound on the Unbiased Statistic and Test\label{sec:boundUnbiasedStat}}

While the previous bounds are of interest since the proof strategy
can be used for general function classes with well behaved Rademacher
averages, a much easier approach may be used 
directly on the unbiased statistic $\mmd_{u}^{2}$ in Lemma \ref{eq:mmd-e-rkhs-unbiased}.
We base our test on the following
theorem, which is a straightforward application of the large deviation
bound on U-statistics of \citet[][p. 25]{Hoeffding63}.
\begin{theorem}\label{thm:hoeffdingQuadraticMMD}
Assume $0\le k(x_{i},x_{j})\le K$, from which it follows $-2K\le h(z_{i},z_{j})\le2K$.
Then
\[
\Pr\left\{ \mmd_{u}^{2}(\Fcal,X,Y)-\mmd^{2}(\Fcal,p,q)>t\right\} \le\exp\left(\frac{-t^{2}m_2}{8 K^{2}}\right)\]
where $m_2 := \lfloor m/2 \rfloor$ (the same bound applies for deviations of $-t$ and below).
\end{theorem}
A consistent statistical test for $p=q$ using $\mmd_{u}^{2}$ is then obtained.
\begin{corollary}\label{cor:HoeffdingThresh}
A hypothesis test of level $\alpha$ for the null hypothesis $p=q$
has the acceptance region $\mmd_{u}^{2} < \left(4K/\sqrt{m}\right)\sqrt{\log(\alpha^{-1})}$.
\end{corollary}
We now compare the thresholds of the two tests. We note first that
the threshold for the biased statistic applies to an estimate of $\mmd$,
whereas that for the unbiased statistic is for an estimate of $\mmd^2$.
Squaring the former threshold to make the two quantities comparable,
the squared threshold in Corollary \ref{lem:firstTest} decreases as $m^{-1}$, whereas the threshold
 in Corollary \ref{cor:HoeffdingThresh}
decreases as $m^{-1/2}$. Thus for sufficiently large\footnote{In the case of $\alpha=0.05$, this is $m\ge 12$.} $m$, the McDiarmid-based threshold will be lower
(and the associated test statistic is in any case biased upwards), and its Type II error will be better
for a given Type I bound.
This is confirmed in our Section \ref{sec:experiments} experiments.
Note, however, that the rate of convergence of the squared, biased MMD estimate to 
its population value remains at $1/\sqrt{m}$ (bearing in mind we take the square of a biased 
estimate, where the bias term decays as $1/\sqrt{m}$).
 

Finally, we note that the bounds we obtained here are rather conservative for a
number of reasons: first, they do not take the actual distributions
into account. In fact, they are finite sample size, distribution free
bounds  that hold even in the worst case scenario.
The bounds could be tightened using 
localization, moments of the distribution, etc. Any such improvements
could be plugged straight into Theorem~\ref{th:general} for a tighter
bound. See e.g.\ \cite{BouBouLug05} for a detailed discussion of recent
uniform convergence bounding methods. 
Second, in
computing \emph{bounds} rather than trying to characterize the
distribution of $\mmd(\Fcal, X,Y)$ explicitly, we force
 our test to be conservative by design. In the following we aim for
an exact characterization of the asymptotic distribution of $\mmd(\Fcal,
X, Y)$ instead of a bound. While this will not satisfy the uniform
convergence requirements, it leads to  superior tests in practice.

\section{Test Based on the Asymptotic Distribution of the Unbiased Statistic}\label{sec:asymptoticTest}

We now propose a third test, which is based on the asymptotic distribution of
the unbiased estimate of $\mmd^2$ in Lemma \ref{lem:rkhs-mmd}.

\begin{figure}[bt]
\includegraphics[width=0.45\columnwidth]{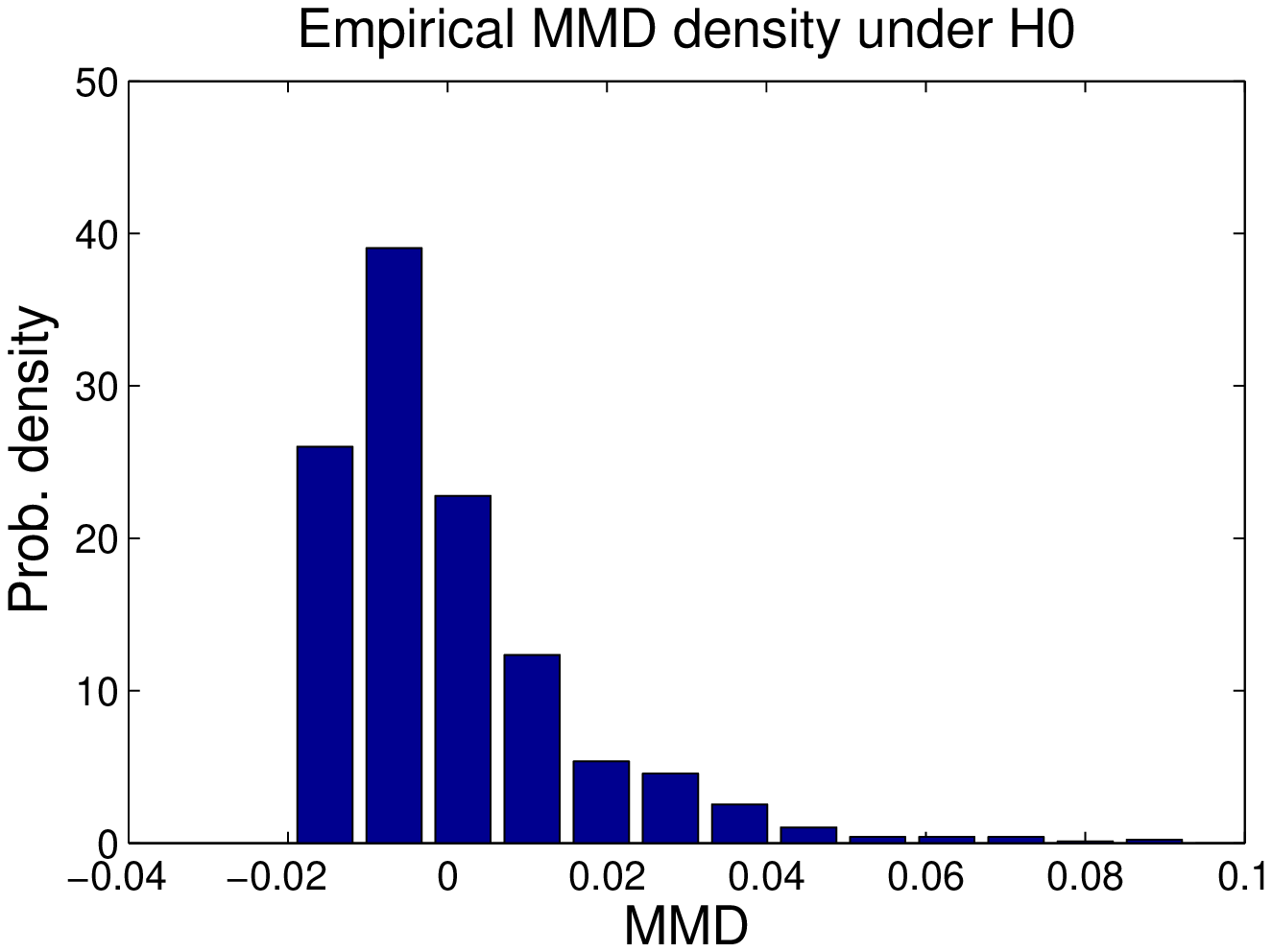} 
\includegraphics[width=0.45\columnwidth]{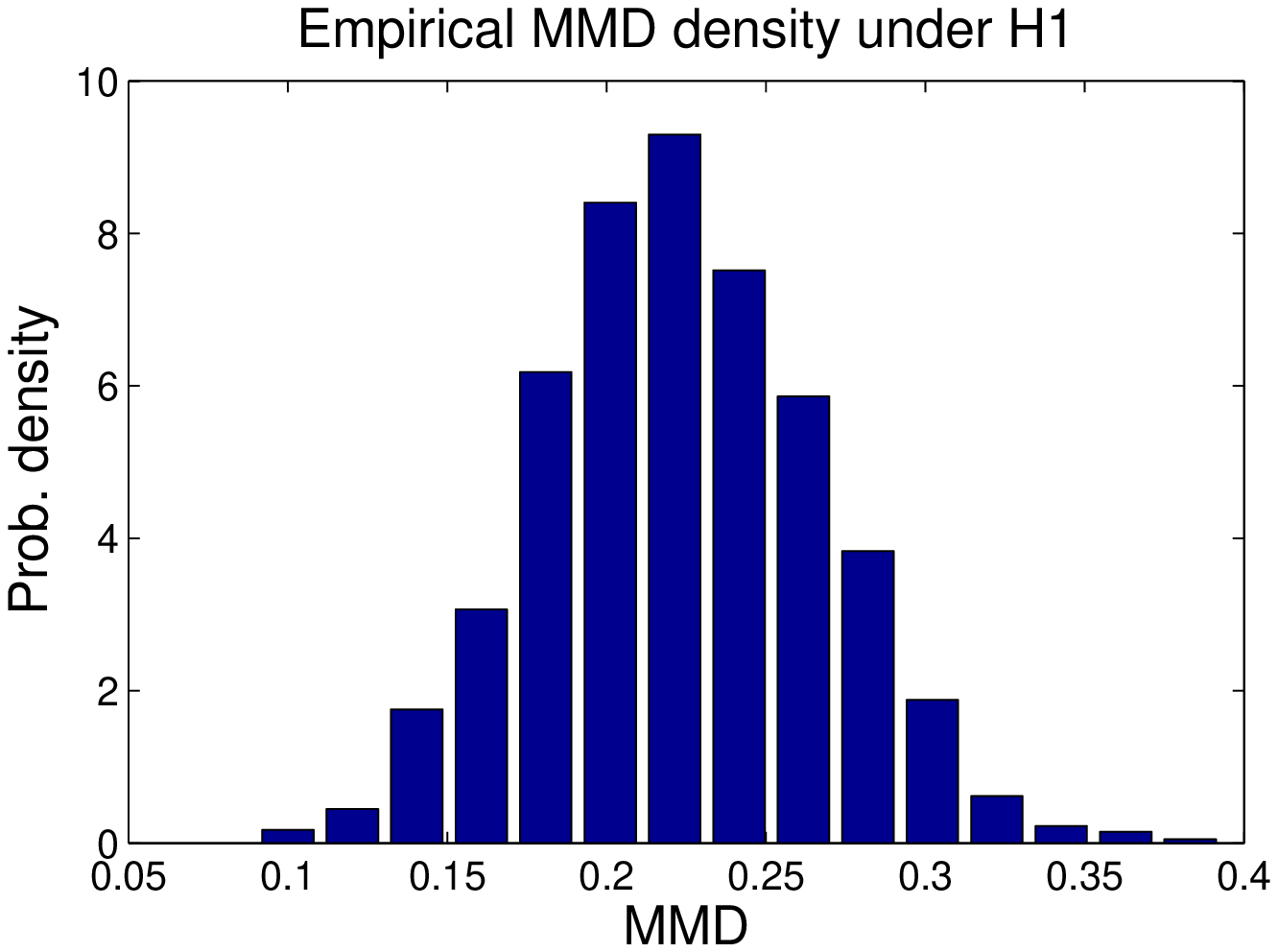} 
\caption{{\bf Left:} Empirical distribution of the MMD under $\Hcal_0$, with $p$ and $q$
both Gaussians with unit standard deviation, using 50 samples from each. {\bf Right:} Empirical distribution of the MMD under $\Hcal_1$, with $p$  a Laplace distribution with unit standard deviation, and $q$ a Laplace distribution with standard deviation $3 \sqrt{2}$, using 100 samples from each.
In both cases, the histograms were obtained by computing 2000
independent instances of the MMD.}
\label{fg:distributionOfMMD}
\end{figure}

\begin{theorem}\label{theorem:hoeffdingNormal}
We assume $\Eb\left( h^2 \right)<\infty$. Under $\Hcal_1$,   $\mmd_u^2$ converges in distribution \citep[see e.g.][Section 7.2]{GriSti01} 
to a Gaussian according to
$$
m^{\frac{1}{2}}\left( \mmd^2_u  - \mmd^2\sbr{\Fcal,p,q} \right)
\overset{D}{\rightarrow}
\mathcal{N}\left(0, \sigma^2_u \right),
\:
$$
where $\sigma^2_u = 4
\left(
\Eb_{z}
\left[
 (\Eb_{z'} h(z,z') )^2
\right]
 - \left[ \Eb_{z,z'} ( h(z,z') ) \right]^2
\right)$, uniformly at rate $1/\sqrt{m}$  \cite[Theorem B, p. 193]{Serfling80}. 
Under $\Hcal_0$, the U-statistic is degenerate, meaning $\Eb_{z'}h(z,z')=0$. In this case, $\mmd_u^2$ converges in distribution according to
\begin{equation}\label{eq:MMD_under_H0}
m \mmd_u^2
\overset{D}{\rightarrow}
\sum_{l=1}^{\infty}
\lambda_l \left[
z_l^2 - 2
\right],
\end{equation}
where $z_l\sim \mathcal{N}(0,2)$  i.i.d., $\lambda_i$ are the solutions to the eigenvalue equation
$$
\int_{\mathcal{X}}
\tilde{k}(x,x')
\psi_{i}(x)dp(x)=\lambda_{i}\psi_{i}(x'),
$$
and  $\tilde{k}(x_{i},x_{j}):=k(x_{i},x_{j})-\Eb_{x}k(x_{i},x)-\Eb_{x}k(x,x_{j})+\Eb_{x,x'}k(x,x')$ is the centred RKHS kernel.
\end{theorem}
The asymptotic
distribution of the test statistic 
under $\Hcal_1$ is given by \citet[][Section 5.5.1]{Serfling80}, and the
distribution under $\Hcal_0$ follows \citet[][Section
5.5.2]{Serfling80} and \citet[][Appendix]{AndHalTit94}; see Appendix
\ref{sec:distribH0} for
details.  
We illustrate the MMD density under both the null and alternative hypotheses by approximating it empirically for both $p=q$ and $p\neq q$. Results are plotted in Figure \ref{fg:distributionOfMMD}.

Our goal is to determine whether the empirical test statistic $\mmd_u^2$ is so large as to be outside the $1-\alpha$ quantile of the null distribution in (\ref{eq:MMD_under_H0}) (consistency of the resulting test is guaranteed by the form of the distribution under $\Hcal_1$). 
One way to estimate this quantile is using the bootstrap  on the aggregated data, following \citet{ArcGin92}.
Alternatively, we may approximate the null distribution by fitting
Pearson curves to its first four moments \cite[Section
18.8]{JohKotBal94}. Taking advantage of the degeneracy of the
U-statistic, we obtain (see Appendix \ref{sec:momentsH0}) 
\begin{align}
\Eb\left(\left[\mmd_u^2\right]^2\right) & 
= \frac{2}{m(m-1)}\Eb_{z,z'}\left[h^{2}(z,z') \right] 
\text{ and } \\
\label{moment3}
\Eb\left(\left[\mmd_u^2\right]^3\right) & 
=
\frac{8(m-2)}{m^{2}(m-1)^{2}}\Eb_{z,z'}\left[h(z,z')\Eb_{z''}\left(h(z,z'')h(z',z'')\right)\right]
+ O(m^{-4}). 
\end{align}
The fourth  moment $\Eb\left(\left[\mmd_u^2\right]^4\right)$ is not
computed, since it is both very small, $O(m^{-4})$, and expensive to
calculate, $O(m^4)$. Instead, we replace the kurtosis\footnote{The kurtosis is defined in
terms of the fourth and second moments as $\mathrm{kurt}\left(\mmd^2_u\right)
=
\frac{\Eb\left(\left[\mmd_u^2\right]^4\right)}{\left[\Eb\left(\left[\mmd_u^2\right]^2\right)\right]^2} -3$.}
with a lower bound due to \citet{Wilkins44},
 $\mathrm{kurt}\left(\mmd^2_u\right)\ge
\left(\mathrm{skew}\left(\mmd^2_u\right)\right)^2+1$. 



Note that  $\mmd^2_u$ may be negative, since it
is an unbiased estimator of $(\mmd[\Fcal,p,q])^2$. However, the only
terms missing to ensure nonnegativity are the terms $h(z_i, z_i)$, which
were removed to remove spurious correlations between
observations. Consequently we have the bound
\begin{align}
  \mmd_u^2 + \frac{1}{m(m-1)} \sum_{i=1}^m k(x_i, x_i) + k(y_i, y_i)
  - 2k(x_i, y_i) \geq 0.
\end{align}






\section{A Linear Time Statistic and Test\label{sec:linearTimeStatistic}}

While the above tests are already more efficient than the $O(m^2 \log
m)$ and $O(m^3)$ tests described earlier, 
it is still desirable to obtain $O(m)$ tests which do not sacrifice too
much statistical power. Moreover, we would like to obtain tests which
have $O(1)$ storage requirements for computing the test statistic in
order to apply it to data streams. We now describe how to achieve this
by computing the test statistic based on a subsampling of the terms in
the sum.
%
%
%
The empirical estimate in this case is 
obtained by drawing pairs from $X$ and $Y$ respectively \emph{without}
replacement. 
\begin{lemma}
  Recall $m_2 := \lfloor m/2 \rfloor$. The estimator 
  $$\mmd_l^2[\Fcal, X, Y] := \frac{1}{m_2} \sum_{i=1}^{m_2} 
  h((x_{2i-1}, y_{2i-1}), (x_{2i}, y_{2i}))$$
  can be computed in linear time. Moreover, it is an unbiased estimate
  of $\mmd^2[\Fcal, p, q]$. 
\end{lemma}
While it is expected (as we will see explicitly later) that $\mmd_l^2$
has higher variance than $\mmd_u^2$, 
%
it is computationally much
more appealing. In particular, the statistic can be used in stream
computations with need for only $O(1)$ memory, whereas $\mmd_u^2$
requires $O(m)$ storage and $O(m^2)$ time to compute the kernel $h$ on
all interacting pairs. 


Since $\mmd_l^2$ is just the average over a set of random variables,
Hoeffding's bound and the central limit theorem readily allow us to
provide both uniform convergence and asymptotic statements for it with
little effort. The first follows directly from \citet[][Theorem 2]{Hoeffding63}. 
\begin{theorem}\label{cor:asy-linear}
Assume $0\le k(x_i,x_j)\le K$.
Then
\[
\Pr\left\{ \mmd_{l}^{2}(\Fcal,X,Y)-\mmd^{2}(\Fcal,p,q)>t\right\} \le\exp\left(\frac{-t^{2}m_2}{8 K^{2}}\right)\]
where $m_2 := \lfloor m/2 \rfloor$ (the same bound applies for deviations of $-t$ and below).
\end{theorem}
Note that the bound of Theorem~\ref{thm:hoeffdingQuadraticMMD} is identical to
that of Theorem~\ref{cor:asy-linear}, which shows the
former is rather loose. Next we invoke the central limit theorem.

\begin{corollary}
  \label{cor:linearstat}
  Assume $0<\Eb\left( h^2 \right)<\infty$. Then $\mmd_l^2$
  converges in distribution to a Gaussian according to
  $$
  m^{\frac{1}{2}}\left( \mmd^2_l  - \mmd^2\sbr{\Fcal,p,q} \right)
  \overset{D}{\rightarrow}
  \mathcal{N}\left(0, \sigma^2_l \right),
  $$
  where $\sigma^2_l = 2
  \sbr{\Eb_{z,z'} h^2(z,z') - \left[ \Eb_{z,z'} h(z,z') \right]^2}$, 
  uniformly at rate $1/\sqrt{m}$.
\end{corollary}
The factor of $2$ arises since we are averaging over only
$\lfloor m/2\rfloor$ observations. Note the difference in the variance
between Theorem~\ref{theorem:hoeffdingNormal} and Corollary~\ref{cor:linearstat},
namely in the former case we are
interested in the average conditional variance $\Eb_{z}
\var_{z'}[h(z,z')|z]$, whereas in the latter case we compute the full variance
$\var_{z,z'}[h(z,z')]$.

We end by noting another potential approach to reducing the computational cost of
the MMD, by computing a low rank approximation to the Gram matrix
\citep{FinSch01,WilSee01,SmoSch00}.  An incremental computation of the
MMD based on such a low rank approximation would require $O(md)$ storage 
and $O(m d)$ computation (where $d$ is the rank of the approximate Gram 
matrix which is used to factorize \emph{both} matrices) rather than
$O(m)$ storage and $O(m^2)$ operations. That said, it remains to be
determined what effect this approximation would have on the distribution
of the test statistic under $\Hcal_0$, and hence on the test threshold.

\section{Similarity Measures Related to MMD}\label{sec:relatedMethods}

Our main point is to propose a new kernel statistic to test whether
two distributions are the same.  However, it is reassuring to observe
links to other measures of similarity between distributions.

\subsection{Link with $L_2$ Distance between Parzen Window Estimates\label{sec:MMDparzen}}

In this section, we demonstrate the connection between our test statistic
and the Parzen window-based statistic of \citet{AndHalTit94}. We
 show that a two-sample test based on Parzen windows converges
more slowly than an RKHS-based test, also following \citet{AndHalTit94}.
Before proceeding, we motivate this discussion with a short overview
of the Parzen window estimate and its properties \citep{Silverman86}.
We assume a distribution $p$ on $\RR^{d}$, which has an associated
density function also written $p$ to minimise notation. The Parzen
window estimate of this density from an i.i.d. sample $X$ of size
$m$ is 
\begin{align}
  \nonumber
  \hat{p}(x) =\frac{1}{m}\sum_{l=1}^{m}\kappa\left(x_{l}-x\right) 
  \text{ where $\kappa$ satisfies }
  \int_{\mathcal{X}}\kappa\left(x\right)dx =1
  \text{ and } \kappa\left(x\right)\ge0.
  \label{kernelIsNormalised}
\end{align}
We may rescale $\kappa$ according to
$\frac{1}{h_{m}^{d}}\kappa\left(\frac{x}{h_{m}}\right)$.  Consistency of
the Parzen window estimate requires
\begin{equation}
  \lim_{m\rightarrow\infty}h_{m}^{d}=0\quad\mathrm{and}\quad\lim_{m\rightarrow\infty}mh_{m}^{d}=\infty.\label{eq:ParzenDecrease}
\end{equation}
We now show that the $L_2$ distance between Parzen windows density estimates
\citep{AndHalTit94} is a special case of the biased MMD in equation (\ref{eq:mmd-e-rkhs}). Denote by $D_r(p,q) := \nbr{p-q}_r$ the $L_r$
distance.  For $r = 1$
the distance $D_r(p,q)$ is known as the Levy distance \cite[]{Feller71},
and for $r = 2$ we encounter distance measures derived from the Renyi
entropy \cite[]{GokPri02}.

Assume that $\hat{p}$ and $\hat{q}$ are given as kernel density
estimates with kernel $\kappa(x-x')$, that is, $\hat{p}(x) = m^{-1} \sum_i
\kappa(x_i- x)$ and $\hat{q}(y)$ is defined by analogy. In this case
\begin{align}
  D_2(\hat{p},\hat{q})^2 & = \int \sbr{\frac 1m \sum_i \kappa(x_i- z) - \frac 1n \sum_i \kappa(y_i-
    z)}^2 dz & \\
  & = 
  \frac 1{m^2} \sum_{i,j=1}^m k(x_i- x_j) + 
  \frac 1{n^2} \sum_{i,j=1}^n k(y_i- y_j) 
   - \frac 2{mn} \sum_{i,j=1}^{m,n} k(x_i- y_j), &
  \label{eq:parzenmeasure} 
\end{align}
where $k(x-y) = \int \kappa(x-z) \kappa(y-z) dz$. By its 
definition $k(x-y)$ is a Mercer kernel \citep{Mercer09}, as it can be
viewed as inner product between $\kappa(x-z)$ and $\kappa(y-z)$ on the domain
$\Xcal$.

A disadvantage of the Parzen window interpretation is that
 when the Parzen window estimates are consistent (which requires the kernel size to decrease with increasing sample size), the resulting two-sample test converges more slowly than using fixed kernels. 
According to \citet[p. 43]{AndHalTit94}, the Type II error of the
two-sample test converges as $m^{-1/2}h_m^{-d/2}$. Thus, given the
schedule for the Parzen window size decrease in (\ref{eq:ParzenDecrease}),
the convergence rate will lie in the open interval $(0,1/2)$: the upper
limit is approached as $h_{m}$ decreases more slowly, and the lower
limit corresponds to $h_{m}$ decreasing near the upper bound of $1/m$.
In other words, by avoiding density estimation, we obtain a better
convergence rate (namely $m^{-1/2}$) than using a Parzen window estimate
with \emph{any} permissible bandwidth decrease schedule.
In addition, the Parzen window interpretation cannot explain the excellent 
performance of MMD based tests in  experimental settings
where the dimensionality greatly exceeds the sample size
(for instance the Gaussian 
toy example in Figure \ref{fig:gaussians}B, for which performance actually improves when the dimensionality increases;
and the microarray datasets in Table \ref{tab:multivariate}).
Finally, our tests are able to employ universal
kernels that cannot be written as inner products between Parzen windows, 
normalized or otherwise:
several examples are given by \citet[Section 3]{Steinwart01b} and \citet[Section 3]{MicXuZha06}.
We may further generalize to kernels on structured objects such as strings and graphs \citep{SchTsuVer04}:
see also our experiments in Section \ref{sec:experiments}.


\subsection{Set Kernels and Kernels Between Probability Measures\label{sec:kernelsBetweenMeasures}}

\cite{GarFlaKowSmo02} propose kernels to deal with sets of observations.
These are then used in the context of Multi-Instance Classification
(MIC). The problem MIC attempts to solve is to find estimators which are
able to infer from the fact that some elements in the set satisfy a
certain property, then the set of observations has this property, too.
For instance, a dish of mushrooms is poisonous if it contains poisonous
mushrooms. Likewise a keyring will open a door if it contains a suitable
key. One is only given the ensemble, however, rather than information
about which instance of the set satisfies the property.

The solution proposed by \cite{GarFlaKowSmo02} is to map the ensembles
$X_i := \cbr{x_{i1}, \ldots, x_{im_i}}$,
where $i$ is the ensemble index and $m_i$ the number of elements in the $i$th ensemble,
 jointly into feature space via
\begin{align}
  \phi(X_i) := \frac{1}{m_i} \sum_{j=1}^{m_i} \phi(x_{ij}),
\end{align}
and use the latter as the basis for a kernel method. This simple
approach affords rather good performance. With the benefit of hindsight,
it is now understandable why the kernel
\begin{align}
  k(X_i, X_j) = \frac{1}{m_i m_j} \sum_{u,v}^{m_i, m_j} k(x_{iu}, x_{jv})
\end{align}
produces useful results: it is simply the kernel between the empirical
means in feature space $\inner{\mu(X_i)}{\mu(X_j)}$ \cite[Eq. 4]{HeiLalBou04}.
 \cite{JebKon03}
later extended this setting by smoothing the empirical densities before
computing inner products. 

Note, however, that property testing for distributions is probably not
optimal when using the mean $\mu[p]$ (or $\mu[X]$ respectively): we are
only interested in determining whether \emph{some} instances in the domain 
have the desired property, rather than making a statement regarding the
distribution of those instances. Taking this into account leads to an
improved algorithm \cite[]{AndTsoHof03}.

\subsection{Kernel Measures of Independence\label{sec:indepMeasures}}

We next demonstrate the application of MMD in determining  whether
two random variables $x$ and $y$ are independent. In other words, assume that pairs of
random variables $(x_i, y_i)$ are jointly drawn from some distribution
$p:=\Pr_{x,y}$. We wish to determine whether this distribution
factorizes, i.e. whether $q:=\Pr_{x}\Pr_{y}$ is the same as $p$.
One application of such an independence measure is in
independent component analysis \citep{Comon94}, where the goal is to find
a linear mapping of the observations $x_i$ to obtain mutually independent outputs.
Kernel methods were employed to solve this problem by \citet{BacJor02,GreBouSmoSch05,GreHerSmoBouetal05}.
In the following we re-derive one
of the above kernel independence measures using  mean operators instead.

We begin by defining
\begin{align*}
  \mu[\Pr_{xy}] & := \Eb_{x,y} \sbr{v((x,y), \cdot)} \\
  \text{and }
  \mu[\Pr_x \times \Pr_y] & := \Eb_{x} \Eb_{y} \sbr{v((x,y), \cdot)}.
\end{align*}
Here we assumed that $\Vcal$ is an RKHS over $\Xcal \times
\Ycal$ with kernel $v((x,y), (x',y'))$. If $x$ and $y$ \emph{are}
dependent, the equality $\mu[\Pr_{xy}] = \mu[\Pr_x \times \Pr_y]$ will
not hold. Hence we may use $\Delta := \nbr{\mu[\Pr_{xy}] - \mu[\Pr_x
  \times \Pr_y]}$ as a measure of dependence.

Now assume that $v((x,y),(x',y')) = k(x,x') l(y,y')$, i.e.\ that the
RKHS $\Vcal$ is a direct product $\Hcal \otimes \Gcal$ of the RKHSs on
$\Xcal$ and $\Ycal$. In this case it is easy to see that
\begin{eqnarray*}
  \Delta^2
 &=&
\nbr{\Eb_{xy} \sbr{k(x,\cdot) l(y,\cdot)} - \Eb_x
    \sbr{k(x,\cdot)} \Eb_y \sbr{l(y,\cdot)}}^2 \\
  &=& \Eb_{xy} \Eb_{x'y'} \sbr{k(x,x') l(y,y')} -
             2 \Eb_x \Eb_y \Eb_{x'y'} \sbr{k(x,x') l(y,y')}  \\
  & & + \Eb_x \Eb_y \Eb_{x'} \Eb_{y'} \sbr{k(x,x') l(y,y')}
\end{eqnarray*}
The latter, however, is exactly what \cite{GreBouSmoSch05} show to
be the Hilbert-Schmidt norm of the cross-covariance operator
between RKHSs: this is zero if and only if $x$ and $y$ are
independent, for universal kernels. We have the following theorem:
\begin{theorem}
  Denote by $C_{xy}$ the covariance operator between random variables
  $x$ and $y$, drawn jointly from $\Pr_{xy}$, where the functions on
  $\Xcal$ and $\Ycal$ are the reproducing kernel Hilbert spaces $\Fcal$
  and $\Gcal$ respectively. Then the Hilbert-Schmidt norm
  $\nbr{C_{xy}}_\mathrm{HS}$ equals $\Delta$.
\end{theorem}

\begin{figure}
\begin{center}
\includegraphics[width=0.7\textwidth]{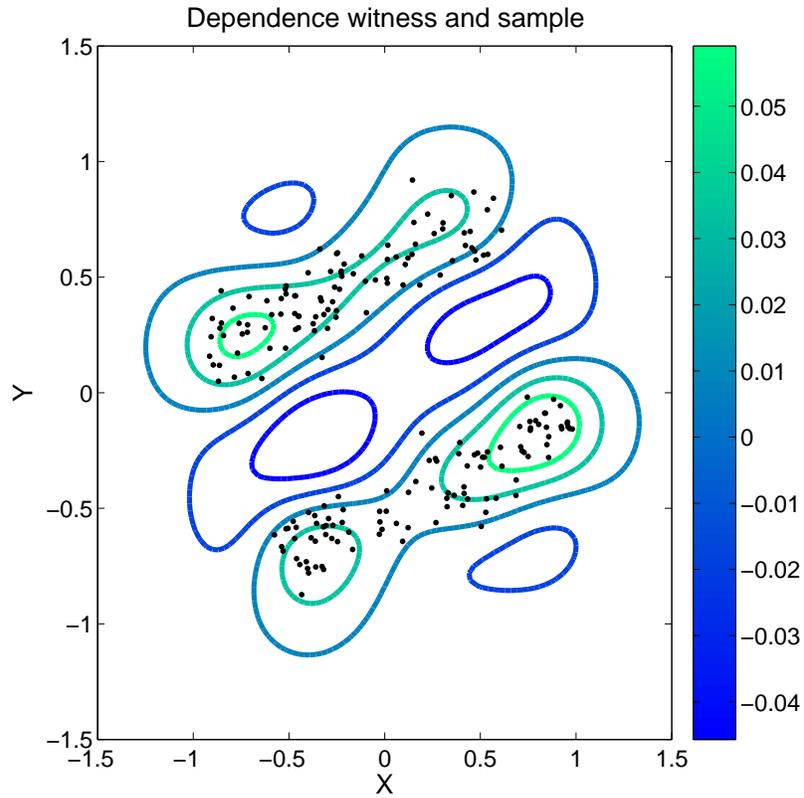}
\end{center}
\caption{Illustration of the function maximizing the mean discrepancy when
  MMD is used as a measure of independence. A sample from dependent random variables $x$ and $y$ is shown in black, and the associated function $f$ that witnesses the MMD is plotted as a contour. The latter was computed empirically on the
  basis of $200$ samples, using a Gaussian kernel with
  $\sigma=0.2$.\label{fig:rotationMapping}}
\end{figure}

Empirical estimates of this quantity are as follows:
\begin{theorem}
  Denote by $K$ and $L$ the kernel matrices on $X$ and $Y$
  respectively, and by $H = I - \one/m$ the projection
  matrix onto the subspace orthogonal to the vector with all entries set
  to $1$. Then $m^{-2} \tr HKHL$ is an estimate of $\Delta^2$ with
 bias  $O(m^{-1})$. With high probability
  the deviation from $\Delta^2$ is $O(m^{-\frac{1}{2}})$.
\end{theorem}
 \cite{GreBouSmoSch05} provide explicit constants. In certain circumstances, including
in the case of RKHSs with Gaussian kernels, the empirical $\Delta^2$ may also
be interpreted in terms of a smoothed difference  between the joint empirical characteristic function (ECF) and the
product of the marginal ECFs \citep{Feuerverger93,Kankainen95}. This interpretation does not hold in
all cases, however, e.g. for kernels on strings, graphs, and other structured spaces.
 An illustration of the
 witness function $f\in\Fcal$ from Definition \ref{def:mmd} is provided in Figure \ref{fig:rotationMapping}. This is a smooth function which has large magnitude where
the joint density is most different from the product of the marginals.

We remark that a hypothesis test based on the above kernel statistic is more complicated
than for the two-sample problem, since the product of the marginal distributions
is in effect simulated by permuting the variables of the original sample. 
Further details are provided by \cite{GreFukTeoSonetal07}.

\subsection{Kernel Statistics Using a Distribution over Witness Functions}

\cite{ShaDol07} define a distance between distributions as follows:
let $\Hcal$ be a set of functions on $\Xcal$ and $r$ be a
probability distribution over $\Fcal$. Then the distance between two
distributions $p$ and $q$ is given by 
\begin{align}
  \label{eq:shado}
  D(p,q) := \Eb_{f \sim r(f)} \abr{\Eb_{x \sim p}[f(x)] - \Eb_{x \sim
      q}[f(x)]}.
\end{align}
That is, we compute the average distance between $p$ and $q$ with
respect to a distribution of test functions. 
\begin{lemma}
  Let $\Hcal$ be a reproducing kernel Hilbert space, $f\in\Hcal$, and
   assume $r(f) = r(\nbr{f}_{\Hcal})$ with finite $\Eb_{f \sim r}[\|f\|_\Hcal]$.
  Then $D(p,q) = C \nbr{\mu[p] - \mu[q]}_{\Hcal}$ for some constant $C$ which
  depends only on $\Hcal$ and $r$.
\end{lemma}
\begin{proof}
By definition $\Eb_{p}[f(x)] = \inner{\mu[p]}{f}_{\Hcal}$.
Using linearity of the inner product, Equation~\eq{eq:shado} equals
\begin{align*}
   & \int \abr{\inner{\mu[p] - \mu[q]}{f}_{\Hcal}} \mathrm{d}r(f)\\
   = & \nbr{\mu[p] - \mu[q]}_{\Hcal} \int \abr{\inner{\frac{\mu[p] - \mu[q]}{\nbr{\mu[p] - \mu[q]}_{\Hcal}}}{f}_{\Hcal}} \mathrm{d}r(f), \end{align*}
where the integral is independent of $p,q$. To see this, note that for any $p,q$, $\frac{\mu[p] - \mu[q]}{\nbr{\mu[p] - \mu[q]}_{\Hcal}}$ is a unit vector which can turned into, say, the first canonical basis vector by a rotation which leaves the integral invariant, bearing in mind that $r$ is rotation invariant.
\end{proof}

\subsection{Outlier Detection\label{sec:relatedProblemsToTwoSample}}

%



An application related to the two sample problem is that of outlier detection:
this is the question of whether a novel point is generated from the same
distribution as a particular i.i.d. sample. In a way, this is a special
case of a two sample test, where the second sample contains only one
observation. Several methods essentially rely on the distance between a
novel point to the sample mean in feature space to detect outliers. 

For instance, \cite{DavGreDouRay02} use a related method to deal with
nonstationary time series. Likewise \citet[][p. 117]{ShaCri04} discuss
how to detect novel observations by using the following reasoning: the
probability of being an outlier is bounded both as a function of the
spread of the points in feature space and the uncertainty in the
empirical feature space mean (as bounded using symmetrisation and
McDiarmid's tail bound).

Instead of using the sample mean and variance, \cite{TaxDui99} estimate
the center and radius of a minimal enclosing sphere for the data, the
advantage being that such bounds can potentially lead to more reliable
tests for \emph{single} observations. \cite{SchPlaShaSmoetal01} show
that the minimal enclosing sphere problem is equivalent to novelty
detection by means of finding a hyperplane separating the data from the
origin, at least in the case of radial basis function kernels.

\section{Experiments}
\label{sec:experiments}


We conducted distribution comparisons using our
MMD-based tests on datasets from three real-world domains: database
applications, bioinformatics, and neurobiology.
We investigated both uniform convergence approaches ($\mmd_b$ 
with the Corollary \ref{lem:firstTest} threshold, and
$\mmd^2_u$ H with the Corollary \ref{cor:HoeffdingThresh} threshold); the asymptotic
approaches with bootstrap ($\mmd^2_u$ B) and moment matching to Pearson curves
 ($\mmd^2_u$ M), both described in Section \ref{sec:asymptoticTest};
and the asymptotic approach using the linear time statistic ($\mmd^2_l$) from Section \ref{sec:linearTimeStatistic}.
We also compared against several alternatives from the literature (where applicable): the multivariate
t-test, the Friedman-Rafsky Kolmogorov-Smirnov generalisation \emph{(Smir)}, the Friedman-Rafsky
Wald-Wolfowitz generalisation \emph{(Wolf)}, the Biau-Gy\"orfi test \emph{(Biau)}, and the Hall-Tajvidi test \emph{(Hall)}. 
See Section \ref{sec:otherTestsReview} for details regarding these tests. Note that we
do not apply the Biau-Gy\"orfi test to high-dimensional problems (since the required space partitioning
is no longer possible), and that MMD is the
only method applicable to structured data such as graphs.

An important issue in the practical application of the MMD-based tests
is the selection of the kernel parameters. We illustrate this with a
Gaussian RBF kernel, where we must choose the kernel width $\sigma$ (we
use this kernel for univariate and multivariate data, but not for
graphs). The empirical MMD is zero both for kernel size $\sigma=0$
(where the aggregate Gram matrix over $X$ and $Y$ is a unit matrix), and
also approaches zero as $\sigma \rightarrow \infty$ (where the aggregate
Gram matrix becomes uniformly constant). We set $\sigma$ to be the
median distance between points in the aggregate sample, as a compromise
between these two extremes: this remains a heuristic, similar to those
described in \cite{TakLeSeaSmo06,Scholkopf97},  and the
optimum choice of kernel size is an ongoing area of research.

\subsection{Toy Example: Two Gaussians}

\begin{figure}
  \centering
  \includegraphics[width=\textwidth]{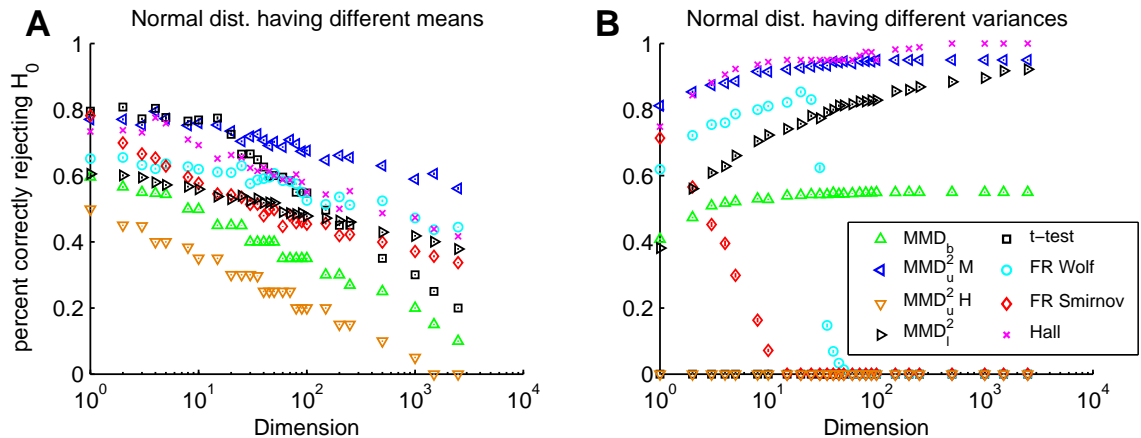}
  \caption{Type II performance of the various tests when separating
two Gaussians, with test level $\alpha=0.05$.
 {\bf A} Gaussians have same variance and different means.
    {\bf B} Gaussians have same mean and different variances.}
  \label{fig:gaussians}
\end{figure}

In our first experiment, we investigated the scaling performance of the various tests
as a function of the dimensionality $d$ of the space $\Xcal\subset \RR^d$, when both $p$ and $q$ were Gaussian.
We considered values of $d$ up to 2500: the performance of the MMD-based tests 
 cannot therefore be explained in the context
of density estimation (as in Section \ref{sec:MMDparzen}), since the associated density estimates are 
necessarily meaningless here.
The levels for all tests were set at $\alpha=0.05$,  $m=250$ samples were used, and results
were averaged over $100$ repetitions.
In the first case, the distributions had different means and
    unit variance.  The  percentage of times the null
    hypothesis was correctly rejected
    over a set of Euclidean distances between 
    the distribution means (20 values logarithmically spaced from 0.05 to 50), 
    was computed as a function of the dimensionality of the normal distributions.  
In case of the t-test, a
    ridge was added to the covariance estimate, to avoid singularity
    (the ratio of largest to smallest eigenvalue was ensured to be at most 2).
In the second case, samples were drawn from distributions
    ${\cal N}(0,\mathbf{I})$ and ${\cal N}(0,\sigma^2\mathbf{I})$ with different variance. The  percentage of
null rejections
    was averaged over 20 $\sigma$ values  logarithmically 
    spaced from $10^{0.01}$ to $10$. The t-test was not compared in this case, since its
output would have been irrelevant.
Results are plotted in Figure \ref{fig:gaussians}.

In the case of  Gaussians with  differing means, we observe the t-test performs best in low dimensions, however
its performance is severely weakened when the number of samples exceeds the number of dimensions. The 
performance of $MMD^2_u$ M is comparable to the t-test for low sample sizes,
and outperforms all other methods for larger sample sizes. The worst performance
is obtained for  $MMD_u^2$ H, though $MMD_b$  also does relatively poorly: this is 
unsurprising given that these tests derive from distribution-free large deviation bounds, 
whereas the sample size is relatively small. Remarkably, $MMD^2_l$ performs quite well compared with classical 
tests in high dimensions.

In the case of Gaussians of differing variance, the \emph{Hall} test performs best, followed
closely by $MMD^2_u$.  \emph{FR Wolf} and (to a much greater extent) \emph{FR Smirnov} 
both have difficulties in high dimensions,  failing completely once the dimensionality becomes too great.
The linear test $MMD^2_l$ again performs surprisingly well, almost matching the $MMD^2_u$ performance 
in the highest dimensionality. Both $MMD_u^2$H and $MMD_b$ perform poorly, the former failing completely:
this is one of several illustrations we will encounter of the much greater tightness of the Corollary \ref{lem:firstTest}
threshold over that in Corollary \ref{cor:HoeffdingThresh}.


\subsection{Data Integration}
In our next application of MMD, we performed distribution testing for
data integration: the objective being to aggregate two datasets into a
single sample, with the understanding that both original samples were generated
from the same  distribution.  Clearly, it is important to
check this last condition before proceeding, or an analysis could
detect patterns in the new dataset that are caused by combining
the two different source distributions, and not by real-world
phenomena.
We chose several real-world settings to perform this task:
we compared microarray data from normal and
tumor tissues (Health status), microarray data from different subtypes
of cancer (Subtype), and local field potential (LFP) electrode recordings 
from the Macaque primary visual cortex (V1) with  and without
spike events \citep[Neural Data I and II, as described in more detail by][]{RasGreMurMasetal08}. In all cases, the two data sets have
different statistical properties, but the detection of these differences
is made difficult by the high  data dimensionality (indeed, for the microarray data,
density estimation is impossible given the sample size and data dimensionality, and no
successful test can rely on accurate density estimates as an intermediate step). 

We applied our tests to these datasets in the following fashion. Given
two datasets A and B, we either chose one sample from A and the other
from B \emph{(attributes = different)}; or both samples from either A
or B \emph{(attributes = same)}. We then repeated this process up to
1200 times.  Results are reported in Table~\ref{tab:multivariate}. 
Our asymptotic tests perform better than all competitors besides \emph{Wolf}: in
the latter case, we have greater
Type II error for one neural dataset, lower Type II error on the Health Status data (which has
very high dimension and low sample size),
and identical (error-free) performance on the remaining examples. We note that the 
Type I error of the bootstrap test on the Subtype dataset is far from its design value of
$0.05$, indicating that the Pearson curves provide a better threshold estimate for these low sample sizes. 
For the remaining datasets, the Type I errors of the Pearson and Bootstrap approximations are close.
Thus, for larger datasets, the bootstrap is to be preferred, since it costs $O(m^2)$, compared
with a cost of $O(m^3)$ for Pearson (due to the cost of computing (\ref{moment3})). Finally, the uniform convergence-based tests 
are too conservative, with $\mathrm{MMD}_b$ finding differences in distribution only for the data with largest sample size,
and $\mathrm{MMD}^2_u$ H never finding differences.

\begin{table}[h]
\centering
\small
 \begin{tabular}{|l|l|r|r|r|r|r|r|r|r|r|r|} \hline
Dataset & Attr.  & $\mathrm{MMD}_b$ &   $\mathrm{MMD}^2_u$ H & $\mathrm{MMD}^2_u$  B& $\mathrm{MMD}^2_u$ M& t-test           & Wolf & Smir & Hall \\\hline\hline  
Neural Data I  & Same           & 100.0  & 100.0 & 96.5  & 96.5 & 100.0   &97.0    &95.0       &96.0  \\\hline 
               & Different      & 38.0   & 100.0 & \textbf{0.0} & \textbf{0.0}  & 42.0   &\textbf{0.0}     &10.0       &49.0  \\\hline\hline 

Neural Data II &  Same           &100.0 & 100.0 &94.6 &95.2 &100.0  &95.0     &94.5       &96.0   \\\hline
               &  Different      &99.7  & 100.0 &3.3 &3.4$$  &100.0  &\textbf{0.8}      &31.8       &5.9  \\\hline\hline

Health status  &Same          &100.0 &100.0  &95.5  &94.4    &100.0    &94.7  &96.1 &95.6  \\\hline
               &Different     &100.0 &100.0 &1.0 &\textbf{0.8}     &100.0    &2.8   &44.0 &35.7  \\\hline\hline



Subtype & Same &100.0 & 100.0 &99.1 &96.4&100.0& 94.6 &97.3 &96.5  \\ \hline
        &   Different & 100.0 &100.0 &\textbf{0.0}&\textbf{0.0} &100.0 &\textbf{0.0} &28.4 &0.2  \\\hline



\end{tabular}                                                     

\caption{Distribution testing for data integration on multivariate  data. Numbers indicate the
  percentage of repetitions for which the null hypothesis (p=q) was accepted, given $\alpha=0.05$. Sample size
  (dimension; repetitions of experiment): Neural I 4000 (63; 100) ; Neural II 1000 (100; 1200); Health Status 25
  (12,600; 1000); Subtype 25 (2,118; 1000).}
\label{tab:multivariate}
\end{table}

\subsection{Computational Cost}



\begin{figure}
\includegraphics[width=0.45\textwidth]{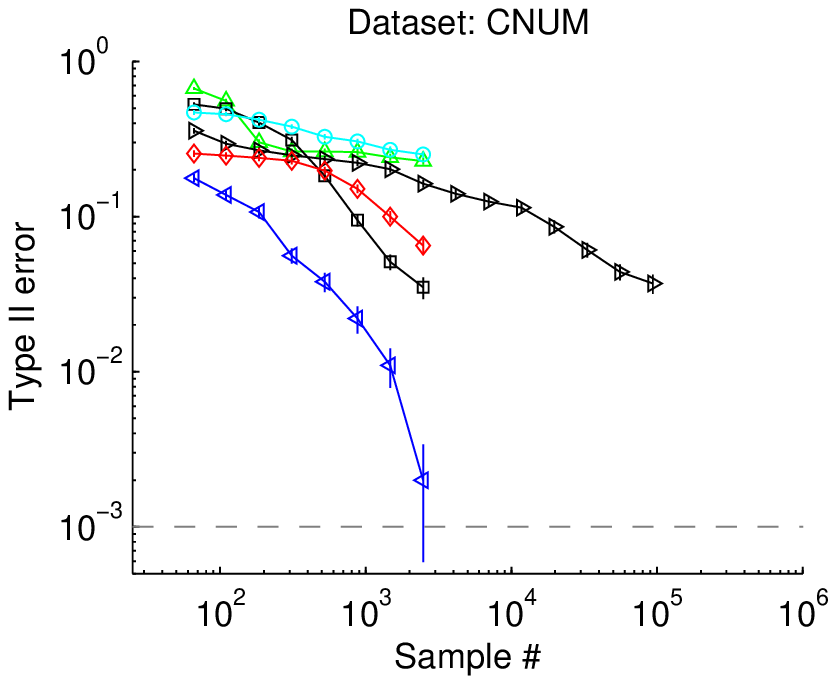}
\includegraphics[width=0.45\textwidth]{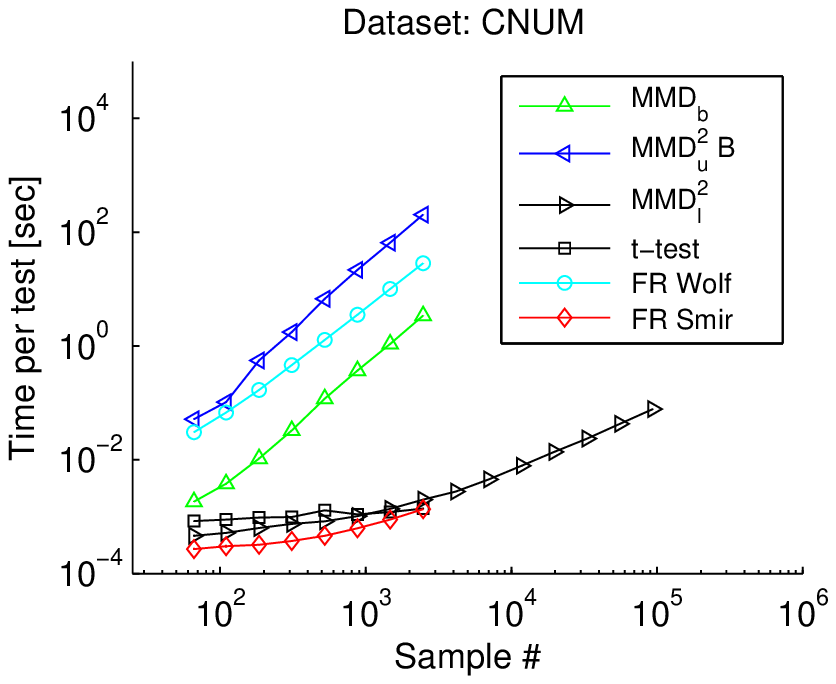}\\
\includegraphics[width=0.45\textwidth]{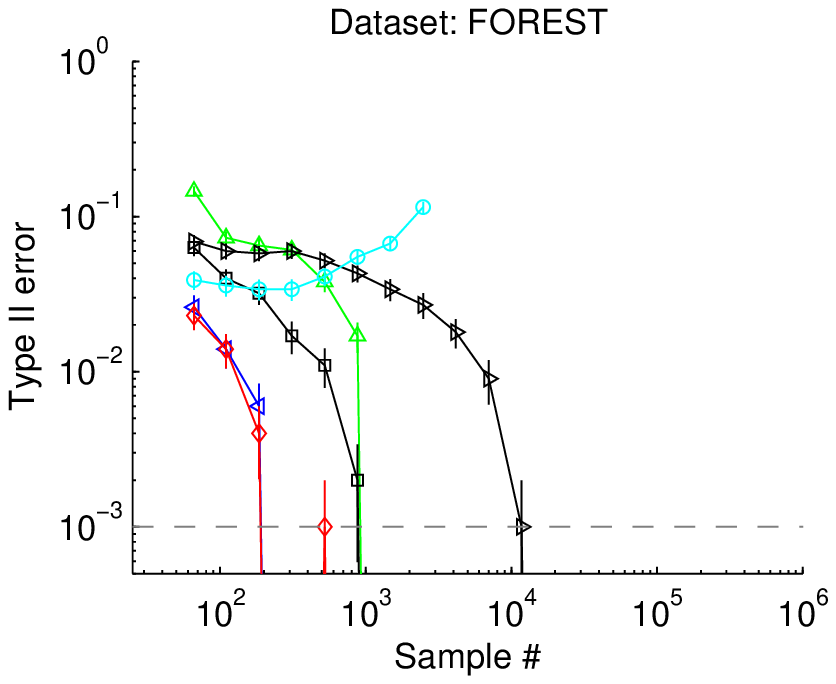}
\includegraphics[width=0.45\textwidth]{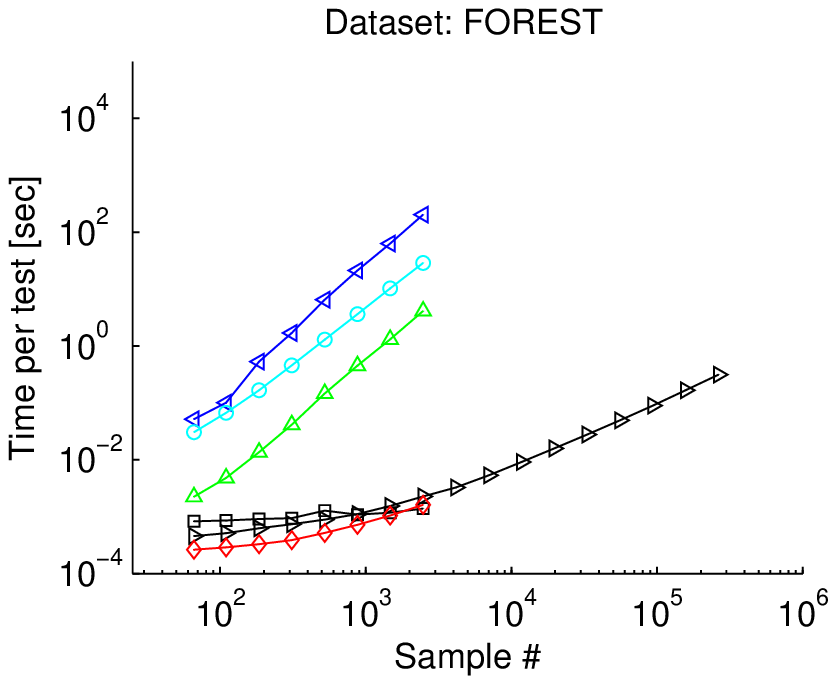}\\
\includegraphics[width=0.45\textwidth]{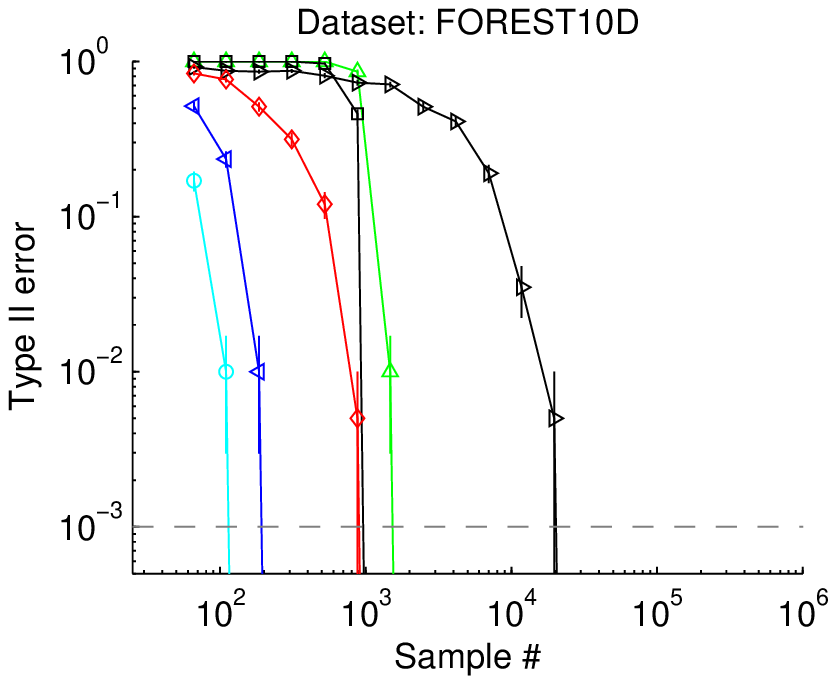}
\includegraphics[width=0.45\textwidth]{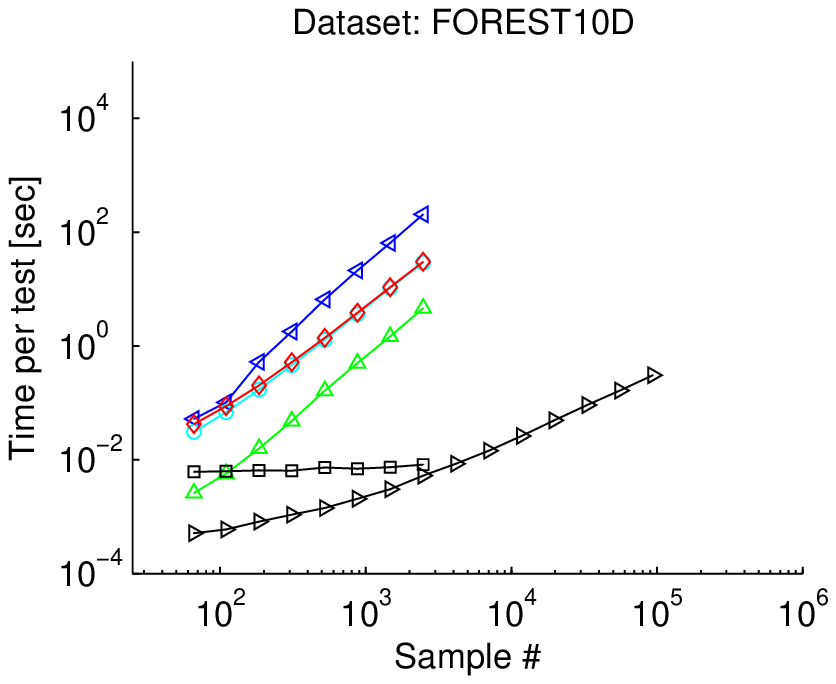}\\
\includegraphics[width=0.45\textwidth]{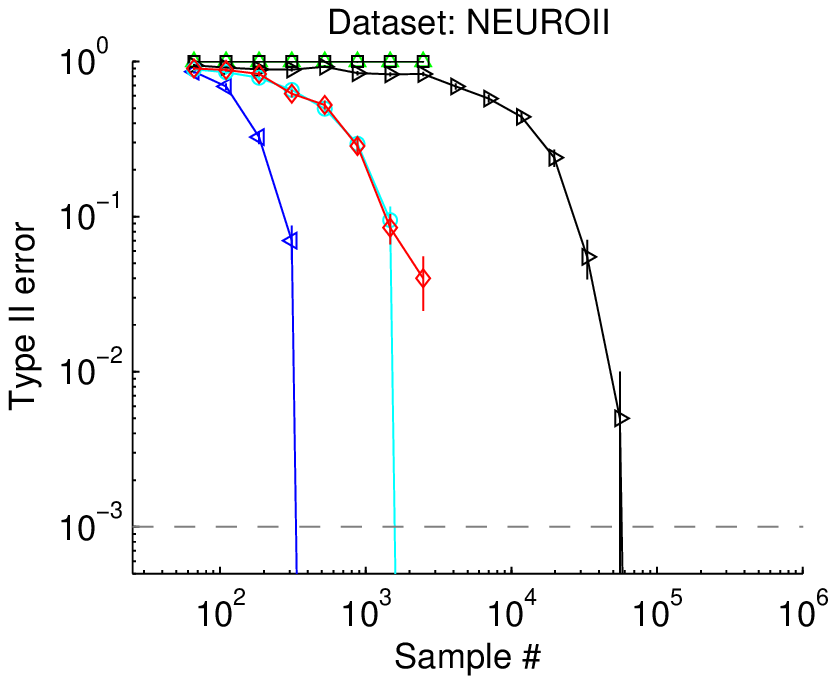}
\includegraphics[width=0.45\textwidth]{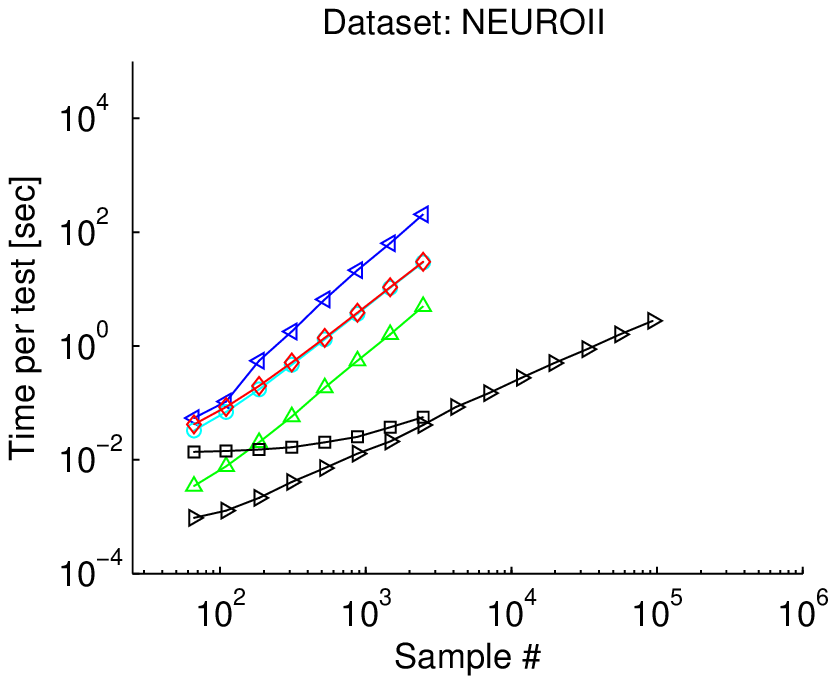}
\caption{Linear vs quadratic MMD. First column is performance, second is runtime.
The dashed grey horizontal line indicates zero Type II error (required due log y-axis)}
\label{fig:linQuadMMD}
\end{figure}

We next investigate the tradeoff between computational cost and performance of the various tests, with particular attention
to how the quadratic time MMD tests from Sections \ref{sec:firstBound} and \ref{sec:asymptoticTest} compare with the linear time MMD-based asymptotic test from Section \ref{sec:linearTimeStatistic}. 
We consider two 1-D datasets (CNUM and FOREST) and two higher-dimensional datasets (FOREST10D and NEUROII).
Results are plotted in Figure \ref{fig:linQuadMMD}. 
If cost is not a factor, then the $\mathrm{MMD}^2_u$ B shows best overall performance
as a function of sample size, with a Type II error dropping to zero as fast or faster than competing 
approaches in three of four cases, and narrowly trailing {\em FR Wolf} in the fourth (FOREST10D).
That said, for datasets CNUM, FOREST, and FOREST10D, the linear time MMD achieves results 
comparable to $\mathrm{MMD}^2_u$ B at a far smaller computational cost, albeit by looking at a great deal more data.
In the CNUM case, however, the linear test is not able to achieve zero error even for the largest
data set size. For the NEUROII data, attaining zero Type II error has about the same cost for both approaches.
The difference in cost of $\mathrm{MMD}^2_u$ B and $\mathrm{MMD}_b$ is due to the bootstrapping required for the former,
which produces a  constant offset in cost between the two (here 150 resamplings were used). 

The $t$-test also performs well in three of the four problems, and in fact represents the best
 cost-performance tradeoff in these three datasets (i.e. while it requires much more data
than $\mathrm{MMD}^2_u$ B for a given level of performance, it costs far less to compute). 
The $t$-test assumes that only the difference in means is important in distinguishing the distributions,
and it requires an accurate estimate of the within-sample covariance;
the test fails completely on the NEUROII data.
We emphasise that the Kolmogorov-Smirnov results in 1-D were obtained using the classical statistic,
and not the Friedman-Rafsky statistic, hence the low computational cost. The cost of both Friedman-Rafsky statistics
is therefore given by the \emph{FR Wolf} cost in this case. 
The latter scales similarly with sample size to the quadratic time MMD
tests, confirming Friedman and Rafsky's observation that obtaining the 
pairwise distances between sample points is the dominant cost of their tests.
We also remark on the unusual behaviour of the Type II error of the 
 \emph{FR Wolf} test in the FOREST dataset, which worsens for increasing sample size.

We conclude that the approach to be recommended when testing homogeneity will depend on the data available: for small amounts of data, the best results are obtained using every observation to maximum effect, and employing the quadratic time $\mathrm{MMD}^2_u$ B test. When large volumes of data are available, a better option is to look at each point only once, which can yield  greater accuracy for a given computational cost. It may also be worth doing 
a t-test first in this case, and only running more sophisticated non-parametric tests if the t-test accepts the null hypothesis, to verify  the distributions are identical in more than just mean.

\subsection{Attribute Matching}
Our final series of experiments addresses automatic attribute matching. Given two
databases, we want to detect corresponding attributes in the schemas of these
databases, based on their data-content (as a simple example, two databases might have respective fields Wage and Salary, 
which are assumed to be observed via a subsampling of a particular population, and we wish to automatically determine that both Wage and Salary denote to the same underlying attribute). 
We use a two-sample test on  pairs of attributes from two databases to
find corresponding pairs.\footnote{Note that corresponding attributes may have
different distributions in real-world databases. Hence, schema matching cannot
solely rely on distribution testing. Advanced approaches to schema matching 
using MMD as one key statistical test are a topic of current research.} 
This procedure is also called {\it table matching} for tables from different databases.
We performed attribute matching as follows:
first, the dataset D  was split into two halves A
and B. Each of the $n$ attributes in A (and B, resp.) was then represented by its
instances in A (resp. B). We then tested all pairs of attributes from A
and from B against each other,  to find the optimal assignment
of attributes ${A_1,\ldots,A_n}$  from A to attributes ${B_1,\ldots,B_n}$ from
$B$. We assumed that A and B contain the same number of attributes.

As a naive approach, one could assume that any possible pair of attributes might
correspond, and thus that every attribute of $A$ needs to be tested against all the attributes of $B$
to find the optimal match.
 We report results for this naive approach, aggregated over all pairs of possible attribute matches,  in
Table~\ref{tab:dataintegration}. We used three datasets: the census income
dataset from the UCI KDD archive (CNUM), the protein homology dataset from
the 2004 KDD Cup (BIO) \citep{CarJoa04}, and the forest dataset from the UCI ML
archive \citep{BlaMer98}. For the final dataset, we performed univariate
matching of attributes (FOREST) and multivariate matching of tables
(FOREST10D) from two different databases, where each table represents
one type of forest. 
Both our asymptotic $\mmd^2_u$-based tests perform as well as or better than the alternatives, notably for CNUM, where the  advantage of $\mmd^2_u$ is  large. Unlike in Table~\ref{tab:multivariate}, the next best alternatives are
not consistently the same across all data: e.g. in BIO they are \emph{Wolf} or \emph{Hall}, whereas in FOREST they are \emph{Smir}, \emph{Biau}, or the t-test. Thus, $\mmd^2_u$ appears to perform more consistently  across the multiple datasets.
The Friedman-Rafsky tests do not always return a Type I error close to the design parameter:  for instance, \emph{Wolf}  has a Type I error of 9.7\% on the BIO dataset (on these data, $\mmd^2_u$ has the joint best Type II error without compromising the designed Type I performance). Finally, $\mathrm{MMD}_b$ performs much better than in Table \ref{tab:multivariate},
although surprisingly it fails to reliably detect differences in FOREST10D. 
The results of $\mathrm{MMD}^2_u$ H are also improved, although it remains among the worst performing methods.

A more principled approach to attribute matching is also possible.
Assume that $\phi(A) = (\phi_1(A_1), \phi_2(A_2), ..., \phi_n(A_n))$: in
other words, the kernel decomposes into kernels on the individual
attributes of A (and also decomposes this way on the attributes of B).
In this case, $MMD^2$ can be written $\sum_{i=1}^{n} \|\mu_i(A_i) -
\mu_i(B_i)\|^2,$ where we sum over the MMD terms on each of the
attributes.  Our goal of optimally assigning attributes from $B$ to
attributes of $A$ via MMD is equivalent to finding the optimal
permutation $\pi$ of attributes of $B$ that minimizes $ \sum_{i=1}^{n}
\|\mu_i(A_i) - \mu_i(B_{\pi(i)})\|^2. $ If we define $C_{ij} =
\|\mu_i(A_i) - \mu_i(B_j)\|^2$, then this is the same as minimizing the
sum over $C_{i, \pi(i)}$. This is the linear assignment problem, which
costs $O(n^3)$ time using the Hungarian method~\citep{Kuhn55}.

While this may appear to be a crude heuristic, it nonetheless defines a
semi-metric on the sample spaces $X$ and $Y$ and the corresponding
distributions $p$ and $q$. This follows from the fact that matching
distances are proper metrics if the matching cost functions are
metrics. We formalize this as follows:
\begin{theorem}
  Let $p, q$ be distributions on $\RR^d$ and denote by $p_i, q_i$ the
  marginal distributions on the $i$-th variable. Moreover, denote by
  $\Pi$ the symmetric group on $\cbr{1, \ldots, d}$. The following
  distance, obtained by optimal coordinate matching, is a semi-metric.
  $$\Delta[\Fcal, p, q] := \min_{\pi \in \Pi} 
  \sum_{i=1}^d \mmd[\Fcal, p_i, q_{\pi(i)}]
  $$
\end{theorem}
\begin{proof}
  Clearly $\Delta[\Fcal, p, q]$ is nonnegative, since all of its
  summands are. Next we show the triangle inequality. Denote by $r$ a
  third distribution on $\RR^d$ and let $\pi_{p,q}, \pi_{q,r}$ and
  $\pi_{p,r}$ be the distance minimizing permutations between $p, q$ and
  $r$ respectively. It then follows that 
  \begin{align*}
    \Delta[\Fcal, p, q] + \Delta[\Fcal, q, r] & = 
    \sum_{i=1}^d \mmd[\Fcal, p_i, q_{\pi_{p,q}(i)}] + 
    \sum_{i=1}^d \mmd[\Fcal, q_i, r_{\pi_{q,r}(i)}] \\
    & \geq \sum_{i=1}^d \mmd[\Fcal, p_i, r_{[\pi_{p,q} \circ \pi_{q,r}](i)}]  
    \geq \Delta[\Fcal, p, r].
  \end{align*}
  Here the first inequality follows from the triangle inequality on
  $\mmd$, that is
  $$\mmd[\Fcal, p_i, q_{\pi_{p,q}(i)}] + \mmd[\Fcal, q_{\pi_{p,q}(i)},
  r_{[\pi_{p,q} \circ \pi_{q,r}](i)}] \geq   
  \mmd[\Fcal, p_i, r_{[\pi_{p,q} \circ \pi_{q,r}](i)}].$$ 
  The second inequality is a result of minimization over $\pi$.  
\end{proof}

\begin{table}[h]
\centering
\small
 \begin{tabular}{|l|l|r|r|r|r|r|r|r|r|r|} \hline

Dataset & Attr.    & $\mathrm{MMD}_b$ & $\mathrm{MMD}^2_u$ H  & $\mathrm{MMD}^2_u$ B& $\mathrm{MMD}^2_u$ M&t-test  &Wolf & Smir & Hall & Biau\\\hline\hline
BIO& Same      &100.0 & 100.0 &93.8 &94.8&95.2    &90.3    &95.8       &95.3 &99.3  \\\hline
   & Different &20.0  & 52.6 &\textbf{17.2}&17.6 &36.2    &\textbf{17.2}    &18.6       &17.9 &42.1  \\\hline\hline



FOREST   & Same     &100.0& 100.0 &96.4 &96.0 &97.4    &94.6    &99.8       &95.5 &100.0  \\\hline
         & Different& 3.9 & 11.0 &\textbf{0.0}   &\textbf{0.0}  &0.2     &3.8     &\textbf{0.0}    &50.1 &\textbf{0.0}  \\\hline\hline

CNUM    & Same      &100.0& 100.0 & 94.5 & 93.8      &94.0  &98.4 &97.5 &91.2 &98.5  \\\hline
        & Different &14.9 & 52.7 & 2.7& \textbf{2.5}&19.17 &22.5 &11.6 &79.1 &50.5  \\\hline\hline

FOREST10D  & Same        &100.0 & 100.0 &94.0 &94.0 &100.0 &93.5    &96.5    &  97.0  &100.0 \\\hline
           & Different   &86.6 &100.0 &\textbf{0.0} &\textbf{0.0}   &\textbf{0.0}    &\textbf{0.0} &1.0          &  72.0  & 100.0  \\\hline

 \end{tabular}

 \caption{Naive  attribute matching on univariate (BIO,
   FOREST, CNUM) and multivariate data (FOREST10D). Numbers indicate
   the percentage of accepted null hypothesis
   (p=q) pooled over attributes. $\alpha=0.05$.  Sample size (dimension; attributes;
   repetitions of experiment): BIO 377 (1; 6; 100); FOREST 538 (1; 10;
   100); CNUM 386 (1; 13; 100); FOREST10D 1000 (10; 2; 100).}
\label{tab:dataintegration}
\end{table}

We tested this 'Hungarian approach' to attribute matching via
$\mathrm{MMD}^2_u$ B on three univariate datasets (BIO, CNUM, FOREST)
and for table matching on a fourth (FOREST10D). To study
$\mathrm{MMD}^2_u$ B on structured data, we obtained two datasets of
protein graphs (PROTEINS and ENZYMES) and used the graph kernel for
proteins from~\cite{BorOngSchVisetal05b} for table matching via the
Hungarian method (the other tests were not applicable to this graph
data). The challenge here is to match tables representing one
functional class of proteins (or enzymes) from dataset A to the
corresponding tables (functional classes) in B. Results  are shown in Table~\ref{tab:structured}. Besides
on the BIO and CNUM datasets, $\mathrm{MMD}^2_u$ B made no errors.

\begin{table}[h]
\begin{tabular}{|l|l|c|c|c|c|}
\hline
Dataset    & Data type   & No. attributes &  Sample size & Repetitions &
\% correct matches\\ \hline\hline
BIO        & univariate   &   6  & 377   & 100  & 90.0   \\\hline 
CNUM       & univariate   &   13 & 386   & 100   & 99.8  \\\hline 
FOREST     & univariate   &   10 & 538   & 100  & 100.0  \\\hline 
FOREST10D  & multivariate &   2 & 1000  & 100  & 100.0  \\\hline 
ENZYME   & structured     &   6   &  50   & 50   & 100.0  \\\hline
PROTEINS   & structured   &   2 & 200   & 50   & 100.0    \\\hline
\end{tabular}
\caption{Hungarian Method for attribute matching via $\mathrm{MMD}^2_u$ B on univariate
  (BIO, CNUM, FOREST), multivariate (FOREST10D), and structured data (ENZYMES, PROTEINS)
  ($\alpha=0.05$; `\% correct matches' is the percentage of the correct attribute
  matches detected over all repetitions). }
\label{tab:structured}
\end{table}


\section{Conclusion}

We have established three simple multivariate tests for comparing two
distributions $p$ and $q$, based on samples of size $m$ and $n$ from
these respective distributions. Our test statistic is 
the maximum mean discrepancy (MMD), defined as the
maximum deviation in the expectation of a function evaluated on each of the random
variables, taken over a sufficiently rich function class: in our case, a universal
reproducing kernel Hilbert space (RKHS). 
Equivalently, the statistic can be written as the norm of the difference between
 distribution feature means in the RKHS. 
We do not
require density estimates as an intermediate step.  Two of
our tests provide Type I error bounds that are exact and
distribution-free for finite sample sizes. We
also give a third test based on quantiles of the asymptotic distribution of the
associated test statistic.  All three tests can be computed in
$O((m+n)^2)$ time, however when sufficient data are available, a linear
time statistic can be used, which employs more data to get better results at smaller computational cost.
In addition, a number of metrics on distributions (Kolmogorov-Smirnov, Earth Mover's,
$L_2$ distance between Parzen window density estimates), as well as certain kernel similarity measures on distributions,
are included within our framework.

While our result establishes that statistical tests based on the MMD
are consistent for universal kernels on compact domains, we draw attention to 
the recent introduction of \emph{characteristic kernels} by \cite{FukGreSunSch08}, these
being kernels for which the mean map is injective. Fukumizu et al. establish
that Gaussian and Laplace kernels are characteristic on $\RR^d$, and thus
the MMD is a consistent test for this domain. 
\citet{Sriperumbuduretal08} further explore the properties of characteristic kernels,
providing a simple condition to determine whether convolution kernels are characteristic, and describing
characteristic kernels  which are not universal on compact domains.
We also note (following Section \ref{sec:kernelsBetweenMeasures}) that the MMD for RKHSs is associated with
 a particular kernel between probability distributions. \citet{HeiLalBou04} describe several
further such kernels, which  induce corresponding distances between feature space distribution mappings:
these may in turn lead to new and  powerful two-sample tests.

Two recent studies have shown that additional divergence measures between distributions
can be obtained empirically through optimization in a reproducing kernel Hilbert space. 
\citet{HarBacMou08} build on the work of \citet{GreBorRasSchSmo07}, 
 considering a homogeneity statistic arising from the kernel Fisher
discriminant, rather than the difference of RKHS means; 
and  \citet{NguWaiJor08} obtain a  KL divergence estimate 
by approximating the ratio of densities (or its log)
with a function in an RKHS.
By design, both these kernel-based statistics 
prioritise different features of
 $p$ and $q$  when measuring the divergence between them, and the resulting effects on
 distinguishability of distributions are therefore of interest.

Finally, we have seen in Section \ref{sec:basicStuffAndReview} that several classical metrics on probability
distributions can be written as maximum mean discrepancies with function classes that are
not Hilbert spaces, but rather Banach, metric, or semi-metric spaces. It would be
of particular interest to establish under what conditions one could write these discrepancies
in terms of norms of differences of mean elements. In particular, \citet{DerLee07} consider
Banach spaces endowed with a semi-inner product, for which a General Riesz Representation
exists for elements in the dual.

\appendix


\section{Large Deviation Bounds for Tests with Finite Sample Guarantees}\label{sec:largeDevBonds}

\subsection{Preliminary Definitions and Theorems}

We need the following theorem, due to \citet{McDiarmid89}.

\begin{theorem}[McDiarmid's inequality]
 Let $f\,:\,\mathcal{X}^{m}\rightarrow\RR$
be a function such that for all $i\in\{1,\ldots,m\}$, there exist
$c_{i}<\infty$ for which\[
\sup_{\bdatax\in\mathcal{X}^{m},\tilde{x}\in\mathcal{X}}|f(x_{1},\ldots x_{m})-f(x_{1},\ldots x_{i-1},\tilde{x},x_{i+1},\ldots,x_{m})|\le c_{i}.\]
Then for all probability measures $p$ and every $\epsilon>0$,\[
p_{\mathsf{x}^{m}}\left(f(\mathbf{x})-\Eb_{\mathsf{x}^{m}}(f(\mathbf{x}))>t\right)<\exp\left(-\frac{2\epsilon^{2}}{\sum_{i=1}^{m}c_{i}^{2}}\right).\]
\end{theorem}
We also define the Rademacher average of the function class $\Fcal$ with respect to the $m$-sample $X$.
\begin{definition}[Rademacher average of $\Fcal$ on $X$]\label{def:rademacher}
Let $\Fcal$ be the unit ball in a universal RKHS on the compact domain $\Xcal$,
with kernel bounded according to $0\le k(x,y)\le K$.
Let $X$ be an i.i.d. sample of size $m$ drawn according to a probability measure $p$ on $\Xcal$, and let $\sigma_{i}$ be i.i.d and take values in $\{-1,1\}$
with equal probability.
 We define the Rademacher average \begin{eqnarray*}
R_{m}(\Fcal,X) & := & \Eb_{\sigma}\sup_{f\in\Fcal}\left|\frac{1}{m}\sum_{i=1}^{m}\sigma_{i}f(x_{i})\right|\\
 & \le & \left(K/m\right)^{1/2},\end{eqnarray*}
 where the upper bound is due to \citet[][Lemma 22]{BarMen02}. Similarly, we define
$$
R_{m}(\Fcal,p)  := 
 \Eb_{p,\sigma}\sup_{f\in\Fcal}\left|\frac{1}{m}\sum_{i=1}^{m}\sigma_{i}f(x_{i})\right|.
$$
\end{definition}

\subsection{Bound when $p$ and $q$ May Differ}
\label{sec:pqdiffproof}

We want to show that the absolute difference between $\mmd(\Fcal,p,q)$
and $\mmd_b(\Fcal,X,Y)$ is close to its expected value, independent of the
distributions $p$ and $q$.  To this end, we prove three intermediate
results, which we then combine.  The first result we need is an upper
bound on the absolute difference between $\mmd(\Fcal,p,q)$ and
$\mmd_b(\Fcal,X,Y)$. We have
\begin{eqnarray}
 &  & \left|\mmd(\Fcal,p,q)-\mmd_b(\Fcal,X,Y)\right| \nonumber\\
 & = & \left|\sup_{f\in\Fcal}\left(\Eb_{p}(f)-\Eb_{q}(f)\right)-\sup_{f\in\Fcal}\left(\frac{1}{m}\sum_{i=1}^{m}f(x_{i})-\frac{1}{n}\sum_{i=1}^{n}f(y_{i})\right)\right| \nonumber\\
 & \le & \underset{\Delta(p,q,X,Y)}{\underbrace{\sup_{f\in\Fcal}\left|\Eb_{p}(f)-\Eb_{q}(f)-\frac{1}{m}\sum_{i=1}^{m}f(x_{i})+\frac{1}{n}\sum_{i=1}^{n}f(y_{i})\right|}}. \label{eq:firstStep}
\end{eqnarray}
Second, we provide an upper bound on the difference between $\Delta(p,q,X,Y)$ and its expectation. Changing either of $x_{i}$ or $y_{i}$ in $\Delta(p,q,X,Y)$ results
in changes in magnitude of at most $2K^{1/2}/m$ or $2K^{1/2}/n$, respectively.
We can then apply McDiarmid's theorem, given a denominator in the
exponent of
\[
m\left(2K^{1/2}/m\right)^{2}+n\left(2K^{1/2}/n\right)^{2}=4K\left(\frac{1}{m}+\frac{1}{n}\right)=4K\frac{m+n}{mn},
\]
to obtain
\begin{equation}\label{eq:step2}
\Pr\left(\Delta(p,q,X,Y)-\Eb_{X,Y}\left[\Delta(p,q,X,Y)\right]>\epsilon\right)\le\exp\left(-\frac{\epsilon^{2}mn}{2K(m+n)}\right).
\end{equation}
For our final  result, we exploit symmetrisation, following e.g. \citet[p. 108]{VaaWel96}, to upper bound the expectation of $\Delta(p,q,X,Y)$. Denoting by $X'$ an i.i.d sample of size $m$ drawn independently of $X$ (and likewise for $Y'$), we have
\begin{eqnarray}
 &  & \Eb_{X,Y}\left[\Delta(p,q,X,Y)\right]\nonumber\\
 & = & \Eb_{X,Y}\sup_{f\in\Fcal}\left|\Eb_{p}(f)-\frac{1}{m}\sum_{i=1}^{m}f(x_{i})-\Eb_{q}(f)+\frac{1}{n}\sum_{i=1}^{n}f(y_{j})\right|\nonumber\\
 & = & \Eb_{X,Y}\sup_{f\in\Fcal}\left|\Eb_{X'}\left(\frac{1}{m}\sum_{i=1}^{m}f(x_{i}')\right)-\frac{1}{m}\sum_{i=1}^{m}f(x_{i})-\Eb_{Y'}\left(\frac{1}{n}\sum_{i=1}^{n}f(y_{j}')\right)+\frac{1}{n}\sum_{i=1}^{n}f(y_{j})\right|\nonumber\\
 & \underset{(a)}{\le} & \Eb_{X,Y,X',Y'}\sup_{f\in\Fcal}\left|\frac{1}{m}\sum_{i=1}^{m}f(x_{i}')-\frac{1}{m}\sum_{i=1}^{m}f(x_{i})-\frac{1}{n}\sum_{i=1}^{n}f(y_{j}')+\frac{1}{n}\sum_{i=1}^{n}f(y_{j})\right|\nonumber\\
 & = & \Eb_{X,Y,X',Y',\sigma,\sigma'}\sup_{f\in\Fcal}\left|\frac{1}{m}\sum_{i=1}^{m}\sigma_{i}\left(f(x_{i}')-f(x_{i})\right)+\frac{1}{n}\sum_{i=1}^{n}\sigma_{i}'\left(f(y_{j}')-f(y_{j})\right)\right|\nonumber\\
 & \underset{(b)}{\le} & \Eb_{X,X'\sigma}\sup_{f\in\Fcal}\left|\frac{1}{m}\sum_{i=1}^{m}\sigma_{i}\left(f(x_{i}')-f(x_{i})\right)\right|+\Eb_{Y,Y'\sigma}\sup_{f\in\Fcal}\left|\frac{1}{n}\sum_{i=1}^{n}\sigma_{i}\left(f(y_{j}')-f(y_{j})\right)\right|\nonumber\\
 & \underset{(c)}{\le} & 2\left[R_{m}(\Fcal,p)+R_{n}(\Fcal,q)\right].\nonumber\\
 & \underset{(d)}{\le} & 2\left[(K/m)^{1/2} + (K/n)^{1/2} \right], \label{eq:step3}
\end{eqnarray}
where (a) uses Jensen's inequality, (b) uses the triangle inequality,
(c) substitutes Definition \ref{def:rademacher} (the Rademacher average), and (d) bounds the Rademacher averages, also via Definition \ref{def:rademacher}.

Having established our preliminary results, we proceed to the proof of Theorem~\ref{th:mmd-diff}.
\begin{proof}[Theorem~\ref{th:mmd-diff}] 
Combining equations (\ref{eq:step2}) and (\ref{eq:step3}), gives
\[
\Pr\left(\Delta(p,q,X,Y)- 2\left[(K/m)^{1/2} + (K/n)^{1/2} \right] >\epsilon\right)\le\exp\left(-\frac{\epsilon^{2}mn}{2K(m+n)}\right).
\]
Substituting  equation (\ref{eq:firstStep}) yields the result.

\end{proof}


\subsection{Bound when $p=q$ and $m=n$}\label{sec:largeDevUnderH0}

In this section, we derive the Theorem \ref{th:mmd-same-a} result, namely the large
deviation bound on the MMD when $p=q$ and $m=n$. Note also that
we consider only positive deviations of $\mmd_b(\Fcal,X,Y)$ from $\mmd(\Fcal,p,q)$,
since negative deviations are irrelevant to our hypothesis test. The
proof follows the same three steps as in the previous section. The
first step in (\ref{eq:firstStep}) becomes\begin{eqnarray}
\mmd_b(\Fcal,X,Y)-\mmd(\Fcal,p,q) & = & \mmd_b(\Fcal,X,X')-0\nonumber \\
 & = & \sup_{f\in\Fcal}\left(\frac{1}{m}\sum_{i=1}^{m}\left(f(x_{i})-f(x_{i}')\right)\right).\label{lalalal}\end{eqnarray}
The McDiarmid bound on the difference between (\ref{lalalal}) and
its expectation is now a function of $2m$ observations in (\ref{lalalal}),
and has a denominator in the exponent of $2m\left(2K^{1/2}/m\right)^{2}=8K/m$.
We use a different strategy in obtaining an upper bound on the expected
(\ref{lalalal}), however: this is now

\begin{eqnarray}
& &\Eb_{X,X'}\left[\sup_{f\in\Fcal}\frac{1}{m}\sum_{i=1}^{m}\left(f(x_{i})-f(x_{i}')\right)\right] \nonumber\\
 & = & \frac{1}{m}\Eb_{X,X'}\left\Vert \sum_{i=1}^{m}\left(\phi(x_{i})-\phi(x_{i}')\right)\right\Vert \nonumber \\
 & = & \frac{1}{m}\Eb_{X,X'}\left[\sum_{i=1}^{m}\sum_{j=1}^{m}\left(k(x_{i},x_{j})+k(x_{i}',x_{j}')-k(x_{i},x'_{j})-k(x_{i}',x_{j})\right)\right]^{\frac{1}{2}}\nonumber \\
 & \le & \frac{1}{m}\left[2m\Eb_{x}k(x,x)+2m(m-1)\Eb_{x,x'}k(x,x')-2m^{2}\Eb_{x,x'}k(x,x')\right]^{\frac{1}{2}}\nonumber\\
 & = & \left[\frac{2}{m}\Eb_{x,x'}\left(k(x,x)-k(x,x')\right)\right]^{\frac{1}{2}}\label{eq:appendBias1}\\
 & \le & \left(2K/m\right)^{1/2}.\label{eq:appendBias2}\end{eqnarray}
We remark that both (\ref{eq:appendBias1}) and (\ref{eq:appendBias2})
bound the amount by which our biased estimate of the population
MMD exceeds zero under $\Hcal_{0}$. Combining the three results,
we find that under $\Hcal_{0}$,\begin{eqnarray*}
p_{X}\left(\mmd_b(\Fcal,X,X')-\left[\frac{2}{m}\Eb_{x,x'}\left(k(x,x)-k(x,x')\right)\right]^{\frac{1}{2}}>\epsilon\right) & < & \exp\left(\frac{-\epsilon^{2}m}{4K}\right)\quad\mathrm{and}\\
p_{X}\left(\mmd_b(\Fcal,X,X')-\left(2K/m\right)^{1/2}>\epsilon\right) & < & \exp\left(\frac{-\epsilon^{2}m}{4K}\right).\end{eqnarray*}



\section{Proofs for Asymptotic Tests} \label{sec:proofs_asymptotic}
We derive results needed in the asymptotic test of Section \ref{sec:asymptoticTest}. Appendix \ref{sec:distribH0} describes the distribution of the empirical MMD under $\Hcal_0$ (both distributions identical). Appendix \ref{sec:momentsH0} contains derivations of the second and third moments of the empirical MMD, also under $\Hcal_0$.

\subsection{Convergence of the Empirical MMD under $\Hcal_0$} \label{sec:distribH0}

We describe the distribution of the unbiased estimator
$\mmd_{u}^{2}[\Fcal,X,Y]$ under the null hypothesis. In this circumstance,
we denote it by $\mmd_{u}^{2}[\Fcal,X,X']$, to emphasise that the second
sample $X'$ is drawn independently from the same distribution as
$X$. We thus obtain the U-statistic\begin{eqnarray}
\mmd_{u}^{2}[\Fcal,X,X'] & = & \frac{1}{m(m-1)}\sum_{i\neq j}k(x_{i},x_{j})+k(x'_{i},x'_{j})-k(x_{i},x'_{j})-k(x_{j},x'_{i})\label{eq:testStat1}\\
 & = & \frac{1}{m(m-1)}\sum_{i\neq j}h(z_{i},z_{j}),\label{eq:testStat2}\end{eqnarray}
where $z_{i}=(x_{i},x_{i}')$. Under the null hypothesis, this U-statistic
is degenerate, meaning\begin{eqnarray*}
\Eb_{z_{j}}h(z_{i},z_{j}) & = & \Eb_{x_{j}}k(x_{i},x_{j})+\Eb_{x_{j}'}k(x'_{i},x'_{j})-\Eb_{x_{j}'}k(x_{i},x_{j}')-\Eb_{x_{j}}k(x_{j},x_{i}')\\
 & = & 0.\end{eqnarray*}
The following theorem from \citet[Section 5.5.2]{Serfling80} then applies.

\begin{theorem}
Assume  $\mmd_{u}^{2}[\Fcal,X,X']$ is as defined in (\ref{eq:testStat2}),
with $\Eb_{z'}h(z,z')=0$, and furthermore assume $0
<\Eb_{z,z'}h^{2}(z,z') < \infty$. Then $\mmd_{u}^{2}[\Fcal,X,X']$ converges in
distribution according to\[
m\mmd_{u}^{2}[\Fcal,X,X']\overset{D}{\rightarrow}\sum_{l=1}^{\infty}\gamma_{l}\left(\chi_{1l}^{2}-1\right),\]
where $\chi_{1l}^{2}$ are independent chi squared random variables
of degree one, and $\gamma_{l}$ are the solutions to the eigenvalue
equation\[
\gamma_{l}\psi_{l}(u)=\int h(u,v)\psi_{l}(v)d\Pr_{v}.\]
\end{theorem}
While this result is adequate for our purposes (since we do not explicitly
use the quantities $\gamma_{l}$ in our subsequent reasoning), it
does not make clear the dependence of the null distribution on the
kernel choice. For this reason, we provide an alternative expression
based on the reasoning of \citet[Appendix]{AndHalTit94}, bearing in
mind the following changes:

\begin{itemize}
\item we do not need to deal with the bias terms $S_{1j}$ seen by \citet[Appendix]{AndHalTit94}
that vanish for large sample sizes, since our statistic is unbiased
(although these bias terms drop faster than the variance);
\item we require greater generality, since we deal with distributions on
compact metric spaces, and not densities on $\RR^{d}$; correspondingly,
our kernels are not necessarily inner products in $L_{2}$ between
probability density functions (although this is a special case).
\end{itemize}
Our first step is to express the kernel $h(z_{i},z_{j})$ of the U-statistic
in terms of an RKHS kernel $\tilde{k}(x_{i},x_{j})$ between feature
space mappings from which the mean has been subtracted, \begin{eqnarray*}
\tilde{k}(x_{i},x_{j}) & := & \left\langle \phi(x_{i})-\mu[p],\phi(x_{j})-\mu[p]\right\rangle \\
 & = & k(x_{i},x_{j})-\Eb_{x}k(x_{i},x)-\Eb_{x}k(x,x_{j})+\Eb_{x,x'}k(x,x').\end{eqnarray*}
The centering terms cancel (the distance between the two points is
unaffected by an identical global shift in both the points), meaning\[
h(z_{i},z_{j})=\tilde{k}(x_{i},x_{j})+\tilde{k}(y_{i},y_{j})-\tilde{k}(x_{i},y_{j})-\tilde{k}(x_{j},y_{i}).\]
Next, we write the kernel $\tilde{k}(x_{i},x_{j})$ in terms of eigenfunctions
$\psi_{i}(x)$ with respect to the probability measure $\Pr_{x}$,\[
\tilde{k}(x,x')=\sum_{l=1}^{\infty}\lambda_{l}\psi_{l}(x)\psi_{l}(x'),\]
where\[
\int_{\mathcal{X}}\tilde{k}(x,x')\psi_{i}(x)d\Pr_{x}(x)=\lambda_{i}\psi_{i}(x')\]
and\begin{equation}
\int_{\mathcal{X}}\psi_{i}(x)\psi_{j}(x)d\Pr_{x}(x)=\delta_{ij}.\label{eq:uncorrelated}\end{equation}


Since \begin{eqnarray*}
\Eb_{x}\tilde{k}(x,v) & = & \Eb_{x}k(x,v)-\Eb_{x,x'}k(x,x')-\Eb_{x}k(x,v)+\Eb_{x,x'}k(x,x')\\
 & = & 0,\end{eqnarray*}
then when $\lambda_{i}\neq0,$ we have that\begin{eqnarray*}
\lambda_{i}\Eb_{x'}\psi_{i}(x') & = & \int_{\mathcal{X}}\Eb_{x'}\tilde{k}(x,x')\psi_{i}(x)d\Pr_{x}(x)\\
 & = & 0,\end{eqnarray*}
and hence \begin{equation}
\Eb_{x}\psi_{i}(x)=0.\label{eq:zeromean}\end{equation}
We now use these results to transform the expression in (\ref{eq:testStat1}).
First, \begin{eqnarray*}
\frac{1}{m}\sum_{i\neq j}\tilde{k}(x_{i},x_{j}) & = & \frac{1}{m}\sum_{i\neq j}\sum_{l=1}^{\infty}\lambda_{l}\psi_{l}(x_{i})\psi_{l}(x_{j})\\
 & = & \frac{1}{m}\sum_{l=1}^{\infty}\lambda_{l}\left(\left(\sum_{i}\psi_{l}(x_{i})\right)^{2}-\sum_{i}\psi_{l}^{2}(x_{i})\right)\\
 & \underset{D}{\rightarrow} & \sum_{l=1}^{\infty}\lambda_{l}(y_{l}^{2}-1),\end{eqnarray*}
where $y_{l}\sim\mathcal{N}(0,1)$ are i.i.d., and the final relation
denotes convergence in distribution, using (\ref{eq:uncorrelated})
and (\ref{eq:zeromean}), following \citet[Section 5.5.2]{Serfling80}. Likewise\begin{eqnarray*}
\frac{1}{m}\sum_{i\neq j}\tilde{k}(x_{i}',x_{j}') & \underset{D}{\rightarrow} & \sum_{l=1}^{\infty}\lambda_{l}(z_{l}^{2}-1),\end{eqnarray*}
where $z_{l}\sim\mathcal{N}(0,1)$, and\begin{eqnarray*}
\frac{1}{m(m-1)}\sum_{i\neq j}\left(\tilde{k}(x_{i},y_{j})+\tilde{k}(x_{j},y_{i})\right) & \underset{D}{\rightarrow} & 2\sum_{l=1}^{\infty}\lambda_{l}y_{l}z_{l}.\end{eqnarray*}
Combining these results, we get
\begin{eqnarray*}
m\mathrm{MMD}^2_u(\Fcal,X,X') & \underset{D}{\rightarrow} & \sum_{l=1}^{\infty}\lambda_{l}\left(y_{l}^{2}+z_{l}^{2}-2-2y_{l}z_{l}\right)\\
 & = & \sum_{l=1}^{\infty}\lambda_{l}\left[(y_{l}-z_{l})^{2}-2\right].
\end{eqnarray*}
Note that $y_{l}-z_{l}$, being the difference of two independent
Gaussian variables, has a normal distribution with mean zero and variance
$2$. This is therefore a quadratic form in a Gaussian random variable
minus an offset $2\sum_{l=1}^{\infty}\lambda_{l}$.

\subsection{Moments of the Empirical MMD Under $\Hcal_0$\label{sec:momentsH0}} 

In this section, we compute the moments of the U-statistic in Section
\ref{sec:asymptoticTest}, under the null hypothesis conditions
\begin{align}
  \Eb_{z,z'}h(z,z')& =0,
  \label{eq:zeroMean2}
\intertext{and, importantly,}
\Eb_{z'}h(z,z')&=0.
\label{eq:degenerateCondition}
\end{align}
Note that the latter implies the former.

\textbf{Variance/2nd moment:} This was derived by \citet[p. 299]{Hoeffding48}, and is
also described by \citet[Lemma A p. 183]{Serfling80}. Applying these results, 
\begin{align*}
 & \Eb\left(\left[\mmd_u^2\right]^2\right) \\
 & =  
\left(\frac{2}{n(n-1)}\right)^{2}\left[\frac{n(n-1)}{2}(n-2)(2)\Eb_{z}\left[(\Eb_{z'}h(z,z'))^{2}\right]+\frac{n(n-1)}{2}\Eb_{z,z'}\left[h^{2}(z,z')\right]\right] \\
 & =  \frac{2(n-2)}{n(n-1)}\Eb_{z}\left[(\Eb_{z'}h(z,z'))^{2}\right]+\frac{2}{n(n-1)}\Eb_{z,z'}\left[h^{2}(z,z')\right]\\
 & =  \frac{2}{n(n-1)}\Eb_{z,z'}\left[h^{2}(z,z')\right],
\end{align*}
where the first term in the penultimate line is zero due to (\ref{eq:degenerateCondition}).
Note that variance and 2nd moment are the same under the zero mean
assumption.

\textbf{3rd moment:} We consider the terms that appear in the expansion
of $\Eb\left(\left[\mmd_u^2\right]^3\right)$. These are all of the form\[
\left(\frac{2}{n(n-1)}\right)^{3}\Eb(h_{ab}h_{cd}h_{ef}),\]
where we shorten $h_{ab}=h(z_{a},z_{b})$, and we know $z_{a}$ and
$z_{b}$ are always independent. Most of the terms
 vanish due to (\ref{eq:zeroMean2})
and (\ref{eq:degenerateCondition}). The first terms that remain take the form
\[
\left(\frac{2}{n(n-1)}\right)^{3}\Eb(h_{ab}h_{bc}h_{ca}),\]
and there are\[
\frac{n(n-1)}{2}(n-2)(2)\]
of them, which gives us the expression
\begin{align}
 & \left(\frac{2}{n(n-1)}\right)^{3}\frac{n(n-1)}{2}(n-2)(2)\Eb_{z,z'}\left[h(z,z')\Eb_{z''}\left(h(z,z'')h(z',z'')\right)\right]\nonumber \\
 & = 
 \frac{8(n-2)}{n^{2}(n-1)^{2}}\Eb_{z,z'}\left[h(z,z')\Eb_{z''}\left(h(z,z'')h(z',z'')\right)\right].\label{eq:pre3rdMoment}
\end{align}
Note the scaling $\frac{8(n-2)}{n^{2}(n-1)^{2}}\sim\frac{1}{n^{3}}$.
The remaining non-zero terms, for which $a=c=e$ and $b=d=f$, take the form
\[
\left(\frac{2}{n(n-1)}\right)^{3}\Eb_{z,z'}\left[h^{3}(z,z')\right],
\]
and there are $\frac{n(n-1)}{2}$ of them, which gives
\begin{equation}
  \left(\frac{2}{n(n-1)}\right)^{2}\Eb_{z,z'}\left[h^{3}(z,z')\right].
  \label{eq:negligible3rdMoment}
\end{equation}
However $\left(\frac{2}{n(n-1)}\right)^{2}\sim n^{-4}$ so this term
is negligible compared with (\ref{eq:pre3rdMoment}). 
Thus, a reasonable approximation to the third moment is
\[
  \Eb\left(\left[\mmd_u^2\right]^3\right)  \approx
  \frac{8(n-2)}{n^{2}(n-1)^{2}}\Eb_{z,z'}\left[h(z,z')\Eb_{z''}\left(h(z,z'')h(z',z'')\right)\right]. 
\]


\paragraph{Acknowledgements:}
We thank Philipp Berens, Olivier Bousquet, John Langford, Omri Guttman,
Matthias Hein, Novi Quadrianto, Le Song, and Vishy Vishwanathan for
constructive discussions; Patrick Warnat (DKFZ, Heidelberg), for providing the microarray datasets;
 and Nikos Logothetis, for providing the neural datasets.
National ICT Australia is funded through the Australian Government's
\emph{Backing Australia's Ability} initiative, in part through the
Australian Research Council.  This work was supported in part by the
IST Programme of the European Community, under the PASCAL Network of
Excellence, IST-2002-506778, and by the Austrian Science Fund (FWF),
project \# S9102-N04.

\end{document}